\documentclass[authoryear, review, times]{elsarticle}
\usepackage[margin=2.5cm]{geometry}
\usepackage{amssymb}
\usepackage{booktabs}
\usepackage{multirow}
\usepackage{color}
\usepackage{lineno}
\usepackage{subfigure}
\usepackage[hidelinks]{hyperref}
\usepackage{graphicx}
\usepackage{xcolor}
\usepackage{soul}
\usepackage{gensymb}

\journal{arXiv}
\begin{document}
\begin{frontmatter}

\title{Advancing the Understanding of Fine-Grained 3D Forest Structures using Digital Cousins and Simulation-to-Reality: Methods and Datasets}

\author[1]{Jing Liu}
\author[2]{Duanchu Wang}
\author[1]{Haoran Gong}
\author[1]{Chongyu Wang}
\author[1]{Jihua Zhu}

\affiliation[1]{organization={School of Software Engineering, Faculty of Electronic and Information Engineering, Xi'an Jiaotong University},
                city={Xi'an},
                postcode={710049}, 
                state={Shaanxi},
                country={China}}

\affiliation[2]{organization={School of Electronic Engineering, Xidian University},
                city={Xi'an},
                postcode={710071}, 
                state={Shaanxi},
                country={China}}
                
\author[1]{Di Wang\corref{cor1}}
\ead{diwang@xjtu.edu.cn}
\cortext[cor1]{Corresponding author}

\begin{abstract}
Understanding and analyzing the spatial semantics and structure of forests is essential for accurate forest resource monitoring and ecosystem research. However, the lack of large-scale and annotated datasets has limited the widespread use of advanced intelligent techniques in this field. To address this challenge, a fully automated synthetic data generation and processing framework based on the concepts of Digital Cousins and Simulation-to-Reality (Sim2Real) is proposed, offering versatility and scalability to any size and platform. Using this process, we created the Boreal3D, the world’s largest forest point cloud dataset. It includes 1000 highly realistic and structurally diverse forest plots across four different platforms, totaling 48,403 trees and over 35.3 billion points. Each point is labeled with semantic, instance, and viewpoint information, while each tree is described with structural parameters such as diameter, crown width, leaf area, and total volume. We designed and conducted extensive experiments to evaluate the potential of Boreal3D in advancing fine-grained 3D forest structure analysis in real-world applications. The results demonstrate that with certain strategies, models pre-trained on synthetic data can significantly improve performance when applied to real forest datasets. Especially, the findings reveal that fine-tuning with only 20\% of real-world data enables the model to achieve performance comparable to models trained exclusively on entire real-world data, highlighting the value and potential of our proposed framework. The Boreal3D dataset, and more broadly, the synthetic data augmentation framework, is poised to become a critical resource for advancing research in large-scale 3D forest scene understanding and structural parameter estimation. 
\end{abstract}

\end{frontmatter}

\section{Introduction}
\label{sec:intro}

Forests cover one-third of the global land area and serve as the world's largest terrestrial carbon sink, storing substantial carbon resources and playing a pivotal role in the global carbon cycle \citep{pan2024enduring}. The evaluation of forest biomass facilitates an in-depth study of the mechanism of the role of the forest carbon sink in the global carbon cycle. Fine-grained three-dimensional (3D) structural information, including tree height, diameter at breast height (DBH), etc., is an important basis for estimating forest biomass \citep{LI2024114322}. Therefore, understanding and analyzing the forest structural information have consistently remained a central research focus in the context of forest biomass estimation and carbon storage assessment \citep{liang2024forestsemantic}.

Light Detection and Ranging (LiDAR) techniques can acquire detailed 3D point cloud data, which provides a comprehensive representation of forest structures. The detailed structures of forest elements, including tree trunks, branches, leaves, as well as the spatial characteristics of understory vegetation and the ground, can be effectively captured. With these advantages, LiDAR technique is widely applied in forestry tasks such as forest inventory \citep{liang2018international}, biomass estimation \citep{BABCOCK20161} and carbon storage assessment \citep{zhao2018utility}. The collection of point cloud data can be carried out through various platforms, such as Satellite Laser Scanning (SLS), Airborne Laser Scanning (ALS), Unmanned Aerial Vehicle based Laser Scanning (ULS), Mobile Laser Scanning (MLS), and Terrestrial Laser Scanning (TLS). Among them, SLS provides global or near-global coverage, but with very sparse ground footprints, limiting its applicability to the assessment of coarse forest information, such as overall heights, across large scales \citep{LIU2021112571}. ALS can achieve rapid scanning of large-scale areas due to its onboard platform, but the resulting point cloud data is relatively sparse. Therefore, it is typically applied in regional to national-scale forest structure assessments \citep{ferraz20123}. ULS collects data based on a drone platform. Compared with ALS, the data acquired by ULS is denser, which enables the study of more detailed individual tree structures \citep{bruggisser2020influence}. MLS and TLS obtain the densest point cloud data from ground. These techniques are commonly selected when fine-grained structural information, such as DBH and crown characteristics, is required \citep{KUKKO2017199, LIANG201663}. However, compared with ALS and ULS, MLS and TLS usually have limited data collection range and efficiency. In general, various platforms exhibit different spatial resolutions and coverage ranges, each presenting unique advantages and limitations that are suitable for diverse tasks in forest research. With the rapid advancement of LiDAR techniques, lightweight and low-cost platforms are continually emerging, revolutionizing our capacity to understand and analyze fine-grained 3D forest structural information at an unprecedented pace \citep{coops2021modelling}.

Over the past two decades, a large number of methods have been developed for 3D forest structural analysis \citep{guo2020lidar}. These approaches typically depend on leveraging prior knowledge of forest structure to design fully automated systems or engineering features to train machine learning models. However, automatically accessing the spatial semantics of forests remains challenging due to the inherent structural complexity and heterogeneity of forest environments. In recent years, advancements in technologies such as deep learning have driven rapid progress in the field of 3D computer vision, including scene perception and understanding, demonstrating remarkable results and significant potential in forest-related applications as well \citep{XIANG2024114078}.

However, despite significant achievements and promising advancements in LiDAR forestry applications, two critical factors continue to impose substantial limitations. The first one is the lack of data annotation. Two fundamental fine-grained forest structure analysis tasks - semantic segmentation, which aims to classify semantic categories (e.g., leaf, wood, understory, or ground), and instance segmentation, which focuses on extracting individual trees from forests - require accurately labeled point clouds to validate the performance of relevant methods. Although LiDAR technology has been extensively utilized in forest structure studies, resulting in the accumulation of vast amounts of point cloud data, the sheer volume and irregular shapes of semantic objects in forest scenes render manual labeling highly labor-intensive and error-prone. This significantly impedes the progress of forest scene understanding. The second one is the difficulty in collecting ground-truth forest structural attributes. While information such as location, DBH, and tree height is routinely collected through forest inventory activities, more complex attributes, including crown-related metrics such as leaf area, tree volume, and biomass, are either too labor-intensive to measure or require destructive methods that harm the trees \citep{du2023lidar}. Consequently, the development and validation of more advanced methods for forest structure assessment across diverse forest types and data platforms necessitate comprehensive datasets with reliable annotations and accurate attribute ground-truth.

In other fields such as urban and transportation, a variety of large-scale datasets have been developed, which successfully supported and promoted the development of relevant tasks \citep{han2024whu}. By contrast, the absence of a unified forest scene dataset has significantly impeded the advancement of 3D visual technologies in the field of forest scene understanding. In recent years, several researchers have recognized this issue, leading to the emergence of notable works such as NEWFOR \citep{eysn2015benchmark}, SYSSIFOSS \citep{weiser2022individual}, FOR-instance \citep{puliti2023instance}, TreeLearn \citep{henrich2024treelearn} and ForestSemantic \citep{liang2024forestsemantic}. However, the scale of these datasets remains limited, and they typically encompass data from a single specific forest type and sensor platform. Currently, the construction of large-scale forest point cloud datasets has become a prominent research focus and an emerging trend. Nonetheless, based on existing dataset works, we identified several key challenges in constructing forest datasets. (1) Geographical constraints. Natural forests, with dense vegetation, are largely inaccessible to humans. Worse still, different types of forests exist in different geographical locations. (2) Ecological concerns. Data collection activities may cause significant disruption to forest ecosystems. (3) High costs. Acquiring multi-sensor point cloud data requires substantial human and material resources, making the costs difficult to control. (4) Labeling challenges. Current methods are insufficient for the efficient and accurate annotation of forest point cloud data. Overall, the first two factors hinder the acquisition of point cloud data from various forest types, while the latter two limit the scope and scale of existing datasets. These factors collectively result in existing datasets potentially falling short of meeting the current demands for forest scene understanding.

In artificial intelligence, digital simulation is frequently employed to generate synthetic data tailored to specific parameters, offering solutions for fields that face common challenges such as data collection difficulties, data gaps, and labor-intensive data annotation. For example, digital twins have emerged as a pivotal direction for the development of modern and future society and industry \citep{9899718}.  However, the complexity of forest environments has significantly constrained the application of digital twin technology in the forestry domain \citep{qiu2023forest}. Unlike digital twins, which require a one-to-one mapping with the real world, the newly proposed concept of digital cousins does not require precise replication of the real world but instead focuses on preserving the key semantics and attributes of the real-world system, providing a more cost-effective and practical approach to representing physical scenes \citep{dai2024automated}. On the other hand, Simulation-to-Reality (Sim2Real), which utilizes knowledge acquired in simulations to enhance the performance of intelligent agents in real-world scenarios, has emerged as a promising strategy in the broader field of artificial intelligence \citep{udupa2024mrfp}. The key motivation and advantage of Sim2Real lie in the fact that information in simulations is fully annotated and possesses known attributes, providing unlimited data and knowledge for models to learn and evolve. However, critical issues, particularly the domain gap between simulation and reality, must be addressed to genuinely enhance real-world applications.

Building on the latest advancements in artificial intelligence, this paper aims to develop an automated and versatile pipeline to generate forest point clouds for any forest type and LiDAR platform. Ultimately, such datasets are intended to facilitate the comprehensive analysis and validation of current methods in forest structure analysis and to drive the development of more advanced techniques, such as deep learning based ones. The main contributions of this paper are as follows:

1. Based on the concept of digital cousins, this paper proposes an automated simulation paradigm for generating forest scenes that closely resemble the real forests. This paradigm, integrated with physics-based LiDAR simulators, is designed to simulate and generate diverse, high-quality point clouds with error-free labels and ground-truth structural attributes. Furthermore, the generation process proposed in this paper can be easily expanded to point cloud data of any platform and any scale.

2. Building upon the proposed automated simulation process, we have developed Boreal3D, the largest forest point cloud dataset to date. This dataset encompasses forest point clouds simulated from multiple sensors for plots with varying structural complexities, supporting a broad spectrum of tasks related to forest structure understanding and attribute estimation.

3. Using the Boreal3D dataset, we conducted a series of comprehensive experiments to validate its effectiveness in enhancing the performance of algorithms across different scenarios, demonstrating its utility for real-world forest scene data. Experimental results confirm that the proposed generation platform can generate reliable point cloud data, and the constructed Boreal3D data significantly improves the model's performance on real forest scene data.

The structure of this paper is as follows. Section \ref{sec:Relatedwork} briefly introduces the current work on forest point cloud datasets and provides an overview of the current research status of semantic and instance segmentation tasks in forest scenes. Section \ref{sec:GenerationWorkflow} describes the working principle and process of the data generation proposed in this paper. Section \ref{sec:Statistics} presents a statistical analysis and visualization of the various attributes of the introduced Boreal3D dataset. Section \ref{sec:Experiment} designs experiments to validate the role of Boreal3D in enhancing real forest point clouds. Section \ref{sec:Discussion} discusses the advantages and challenges of Boreal3D based on the experimental results. Finally, Section \ref{sec:Conclusion} concludes the paper.

\section{Related work}
\label{sec:Relatedwork}
\begin{table}[t]
    \centering
    \resizebox{\columnwidth}{!}{%
    \begin{tabular}{ccccccc}
    \toprule
     & Dataset & Year & Platform &  Plot count &  Tree count & Point count \\
    \midrule
    Real & NEWFOR \citep{eysn2015benchmark} & 2015 & ALS& 18 & - & - \\
        & SYSSIFOSS \citep{weiser2022individual} & 2022& ALS\&ULS\&TLS & 12 & - & - \\
        & FOR-Instance \citep{puliti2023instance} & 2023 & ULS & 33 & 1130 & 0.15B  \\
        & TreeLearn \citep{henrich2024treelearn} & 2024 & MLS & 19 & - & 1.39B  \\
        & ForestSemantic \citep{liang2024forestsemantic} & 2024 & TLS & 6 & 673 & -  \\
    \midrule
    Synthetic & TreePointCloud \citep{lin2020novel} & 2020 &  - & - & - & - \\
        & Smart-Tree \citep{dobbs2023smart} & 2023 & - & - & - & \\
        & LiDAR-Forest \citep{lu2024lidar} & 2024 & - & - & - & - \\
        & TreeNet3D \citep{tang2024treenet3d} & 2024 & - & - & 13000 & - \\
        & \textbf{Boreal3D} & \textbf{2024} & \textbf{ALS\&ULS\&TLS\&MLS} & \textbf{1000} & \textbf{48403} & \textbf{35.38B} \\
    \bottomrule
    \end{tabular}
    
    }
    \caption{
        Statistics of existing forest point cloud datasets. The Boreal3D dataset, constructed in this paper, is highlighted in bold. "-" indicates unknown values. "B" stands for billion.
    }
    \vspace{-0.05in}
    \label{tab:datasets}
\end{table}
\subsection{Real-world Forest Point Cloud Datasets}
Collecting point cloud data from real forest areas is the most common and widely used method for 3D forest scene analysis. These works typically target specific tasks, where point cloud data is collected from selected forest regions, followed by manual annotation or semi-automatic coarse labeling with manual refinement. As shown in Table \ref{tab:datasets}, these studies are typically limited in scale, often focusing on individual trees or encompassing a small number of plots, ranging from a few to several tens. Specifically, Early work by \citet{eysn2015benchmark} created the NEWFOR dataset for research on single-tree detection in alpine regions. This dataset includes ALS point cloud data from 18 plots across eight different regions. SYSSIFOSS \citep{weiser2022individual} includes 12 plots, with simultaneous ALS, ULS, and TLS data collection for each plot. This dataset provides annotations for single-tree segmentation, along with important parameters such as DBH and tree height. The recently introduced FOR-instance dataset \citep{puliti2023instance} is a UAV LiDAR Scanning (ULS) dataset specifically designed for semantic segmentation and instance segmentation tasks. Comprising five distinct subsets with a total of 30 plots, this dataset has effectively facilitated the advancement of deep learning-based methodologies for forest point cloud segmentation \citep{WIELGOSZ2024114367, XIANG2024114078}. Similarly, aimed at promoting deep learning methods for forest point clouds, TreeLearn \citep{henrich2024treelearn} is a manually segmented benchmark forest dataset, consisting of 156 complete trees and 79 partial trees. On the other hand, ForestSemantic \citep{liang2024forestsemantic} is a TLS forest point cloud dataset comprising 6 plots and a total of 669 trees. It provides fine-grained annotations for each tree, including point-wise semantic and instance labels, as well as detailed information on branching orders and topological properties.

\subsection{Synthetic Forest Point Cloud Datasets}
Obtaining point cloud data from real forests can be costly, and the subsequent annotation process further increases the expense of dataset creation. As a result, some researchers have begun exploring the use of simulated data to replace real-world point cloud datasets (Table \ref{tab:datasets}, part Synthetic). \citet{lin2020novel} constructed the TreePointCloud dataset, which provides tree skeleton ground truth for synthetic point clouds. This dataset is derived from complete tree models, with point clouds generated from them, and it includes four types of point clouds produced by introducing noise, random omissions, and other techniques. Furthermore, \citet{dobbs2023smart} aimed to create a large-scale point cloud skeleton estimation benchmark dataset using the SpeedTree software for tree modeling. LiDAR-Forest \citep{lu2024lidar} has developed a programmatic LiDAR simulation process that generates various types of point clouds by adjusting parameters such as scan patterns, beam quantity, and more. TreeNet3D \citep{tang2024treenet3d} is a large-scale synthetic dataset covering 13,000 models of different tree species. It provides multiple labels, including tree structure parameters, branch-leaf separation, and skeleton point annotations.

\subsection{3D Semantic Segmentation in Forest Scenes}
In the context of forest scenes, 3D semantic segmentation is typically concerned with the classification of the primary components of forest elements, including leaves, branches, understory, and ground. Specifically, ground filtering and leaf-wood separation are among the most prevalent semantic segmentation tasks. 

Ground filtering, essential for extracting earth surface points in forest applications like Digital Terrain Model (DTM) generation and tree height normalization, has evolved significantly. Early unsupervised methods relied on geometric priors, using techniques such as sliding window morphological operations \citep{kilian1996capture} and reference surface construction \citep{isprs-archives-XLIII-B2-2020-279-2020}. However, these methods lacked adaptability across diverse environments. Recent supervised approaches, leveraging machine learning and deep learning, have improved generalization. Initial efforts adapted traditional machine learning with manual feature extraction \citep{jahromi2011novel}, while later advancements utilized 2D image segmentation techniques by projecting 3D point clouds into 2D feature maps \citep{GEVAERT2018106}. To preserve spatial structure, 3D CNNs were applied to voxelized point clouds \citep{dai2023deep}, and direct processing of raw point clouds was introduced to eliminate preprocessing \citep{chen2024sc}. These developments have significantly enhanced the efficiency and accuracy of ground filtering.

Branches and leaves are crucial components of forest structure, and their accurate separation is essential for subsequent analyses such as leaf area and aboveground biomass estimation \citep{chen2025impact}. Conventional methods classify points as wood or leaf using geometric features derived from 3D point cloud coordinates and neighborhood relationships \citep{WANG202086}, or laser return intensity (LRI) collected by sensors \citep{calders2022laser}. While geometric features combined with machine learning achieve high accuracy across forest types, LRI is often unstable due to factors like collection distance and vegetation type. Combining multiple features has been shown to improve classification stability \citep{zhu2018foliar}. Additionally, graph-based features extracted from topological networks have proven effective for wood-leaf separation in diverse forest types \citep{tian2022graph}. Recently, deep learning methods have been adapted for direct processing of forest point clouds, achieving significant performance gains \citep{xi2023delineating}. However, these methods face trade-offs that small backbone networks are resource-efficient but may lack feature expression capabilities, while large networks require extensive computational resources and labeled data, limiting their applicability to large-scale forest scenes \citep{arrizza2024terrestrial}.

It should be noted that, recent works have begun to explore the full segmentation of forest semantics \citep{XIANG2024114078}, owing to the development of relevant annotated datasets and deep learning models.

\subsection{3D Instance Segmentation in forest scenes}
Forest 3D instance refers to the individual tree segmentation (ITS) task. ITC requires not only segmenting trees and non-tree parts, but also segmenting trees into individual instances. Crown segmentation based on canopy height model (CHM) is one of the main application scenario for 3D instance segmentation in forest scenes. For example, in CHM, the crowns often have high grayscale values, while the values of tree gaps are relatively low. Therefore, directly applying the watershed segmentation method to CHM can well segment the crowns along the boundaries \citep{li2023individual}. After obtaining the coarse segmentation of the crown through CHM, the local maximum filter is used to locate the treetops, and the region growing algorithm is used to segment the crown \citep{fu2024individual}. The selection of treetops in this method is often affected by the spatial resolution of the point cloud data. The graph-based segmentation algorithm maps the gridded point cloud into a weighted graph, and then uses the direct feature similarity of the sub-graphs to split and segment the crowns \citep{yang2020individual}. However, the process of building the graph requires too much computing resources and is difficult to apply to large-scale point cloud scenes.

Although the CHM method is simple and efficient, it loses a lot of detail information in 3D space, and errors are introduced in the process of calculating CHM. Point-based methods often directly use original point clouds or voxelized point clouds as input to segment tree trunks or crowns, and use clustering methods such as K-means clustering \citep{morsdorf2003clustering}, mean-shift \citep{hu2017adaptive}, voxel space projection \citep{wang2008lidar}, adaptive multi-scale filter \citep{lee2010adaptive}, regional growth method \citep{tao2015segmenting}, etc. to segment single trees. Currently, some researchers apply group-based instance segmentation methods to forest scenes. For example, \citet{isprs-annals-X-1-W1-2023-605-2023} proposed an efficient group-based single tree segmentation method. This method uses random cylinders sampling method for large-scale forest scenes to obtain the input point cloud of the model. The subsequent 3D UNet is used to extract features and connect three sub-branches, namely offset, semantic and embedding. The shifted coordinates of the points predicted as "tree" are then clustered to obtain preliminary instances. These instances are refined and fine-tuned by ScoreNet again to obtain the final instance prediction. The recent work of TreeLearn \citep{henrich2024treelearn} follows the idea of the SoftGroup \citep{vu2022softgroup} to perform instance segmentation in forest point clouds. First, the backbone network is used to predict the offset and semantic segmentation results. Then, the points that are semantically segmented into trees are extracted to obtain the shifted coordinates. Clustering algorithms are used to cluster these points to obtain the final individual tree. 

Overall, point-based methods, particularly those leveraging deep learning, demonstrate significant performance improvements over CHM based approaches. However, several limitations persist. Firstly, these methods, which process raw point clouds directly, require substantial computational resources and suffer from slow processing speeds. Secondly, the accuracy of initial segmentation results critically impacts the performance of subsequent individual tree segmentation tasks.

\section{Generation Workflow of Boreal3D}
\label{sec:GenerationWorkflow}
\begin{figure}
	\centering
	\includegraphics[width=0.7\linewidth]{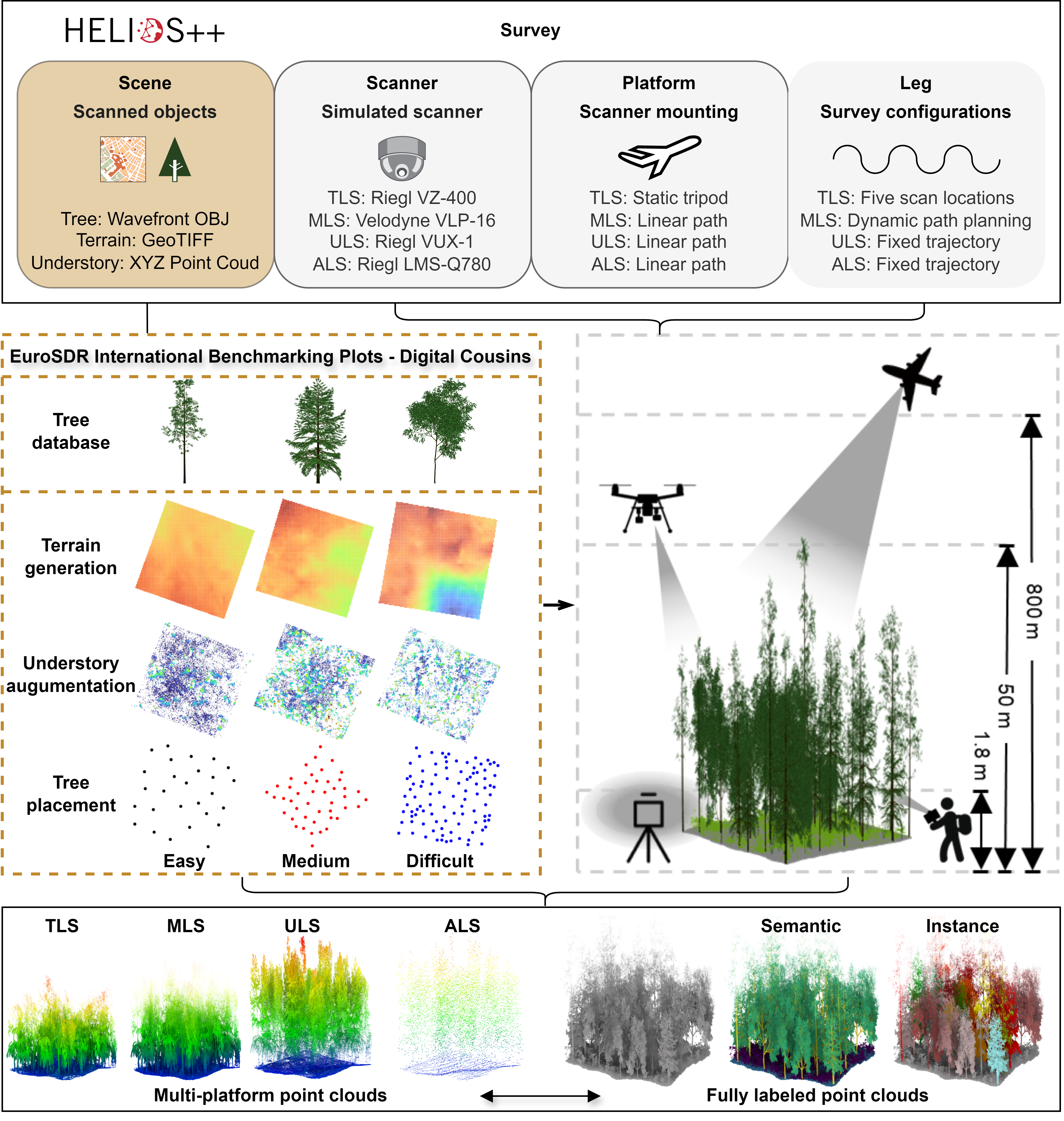}
	\caption{Overview of the workflow of dataset generation.}
    \label{fig:data_construction}
\end{figure}

At the core of our methodology lies the utilization of digital cousins and Simulation-to-Real (Sim2Real) techniques to generate virtual forest plots and corresponding synthetic forest point clouds with ground-truth labels and structural parameters. To this end, we have developed a fully automated pipeline centered around the Helios++ engine, which is a physics-induced and highly versatile LiDAR simulator \citep{heliosPlusPlus}. The pipeline mainly involves two steps: the creation of vertical forest plots and multi-platform point cloud simulation (Figure \ref{fig:data_construction}). The following sections will provide a detailed account of these processes.
\subsection{Helios++ LiDAR Simulator}
In contrast to numerous existing vision-oriented tasks that employ elementary ray tracing techniques or analogous software to simulate point clouds, our emphasis is on large-scale outdoor applications that typically necessitate survey-grade LiDAR measurement. Accordingly, we have elected to use the well-established Helios++ simulator \citep{heliosPlusPlus}. Helios++ is an open-source simulation framework for laser scanning, implemented in the C++ programming language. It considers the physical properties (e.g., laser beam divergence, waveform, and beam scattering) of the laser scanning mechanism in a comprehensive manner and offers a highly versatile modular design that can simulate any kind of laser scanner and mounting platform by specifying their corresponding configurations. Additionally, it is capable of accepting the scene to be scanned in a multitude of formats, including mesh, Geotiff, and point clouds. This renders it particularly well-suited for large-scale and complex scenes, where explicit mesh models are either too challenging to construct or too large in size. Helios++ has been employed extensively in the simulation of forest ecosystems, topographical mapping, and other scientific investigations, thereby substantiating its efficacy \citep{gonzalez2024influence,wang2022plantmove,  weiser2022individual}

The Helios++ system employs the use of Extensible Markup Language (XML) files to define the input elements. The fundamental component is a survey (see Figure \ref{fig:data_construction}). A survey is a controlling file that defines the scene to be scanned, the simulated scanner, and the corresponding platform on which the virtual scanner is mounted and moved through the scene. Furthermore, an additional element, designated as leg, is defined to represent the waypoints for the platform. For detailed technical descriptions and configurations, please refer to https://www.geog.uni-heidelberg.de/gis/helios.html and the GitHub repository https://github.com/3dgeo-heidelberg/helios.git.

\subsection{Virtual Forest Scene Creation}
In order to conduct simulations, it is essential to construct a virtual scene that accurately represents the area of interest. In this study, we concentrate on real-world forest scenes and their fine-grained 3D structures. Accordingly, the virtual forest scene must be realistic and detailed enough to accurately represent the actual environment. However, in contrast to indoor or more regular manmade environments, forests are known to possess a high degree of complexity and heterogeneity in their structural characteristics. This renders the construction process exceedingly challenging.

In this work, we employ the concept of recently proposed digital cousins in the field of artificial intelligence to create virtual forest scenes. In contrast to digital twins, which necessitate the explicit, one-to-one modeling of real-world objects, digital cousins aim to preserve the geometric and semantic characteristics, offering a more feasible and cost-effective representation approach \citep{dai2024automated}. Specifically, we follow the forest plots provided by the EuroSDR international terrestrial laser scanning benchmarking project to guide the creation of virtual forest. The forest plots provided are situated in Evo Finland and encompass a range of forest stand conditions, representing diverse developmental stages, stem densities, and the prevalence of subcanopy growth in boreal forests \citep{liang2018international}. These plots have been classified into three complexity categories: easy, medium, and difficult. The main tree species are Scots pine (\textit{Pinus sylvestris} L.), Norway spruce (\textit{Picea abies} L. Karst.) and silver (\textit{Betula pendula} Roth) and downy (\textit{Betula pubescens} Ehrh.) birches. In total, TLS point clouds of six plots with two for each complexity category are openly available. In this work, we present a fully automatic and versatile pipeline for the generation of virtual forest plots that represent the geometric (e.g., stem density, height distribution) and semantic (e.g., components) attributes of these plots.

\subsubsection{Tree Model Database}
In order to create the digital cousins of these plots, it was first necessary to construct a tree model database. Tree models are constructed using the widely used SpeedTree software (Interactive Data Visualization, Inc. Lexington, SC, USA), which facilitates the creation of highly detailed and realistic tree models. For each tree, the woody component (i.e., the stem and branches) and the foliage (i.e., the leaves) are separated and stored as mesh models. This is done in order to facilitate semantic reasoning in point cloud simulation. To create a database of tree models representing different growth stages, tree models were constructed with heights ranging from 2 to 35 meters, with an increment of one meter. To enhance the diversity of the tree models, three models were created for each height range. This process was repeated for pine, spruce, and birch, resulting in the creation of 306 tree base models in total. These tree models, combined with a range of augmentation techniques (see section \ref{plotcreation}), facilitate the representation of tree diversity while circumventing the necessity of modeling each tree individually, which would otherwise impede the creation of large-scale plots. Furthermore, it is noteworthy that this approach can be extended to encompass a greater number of tree base models.

\subsubsection{Terrain Generation}
The forest plots, which are to be mimicked, are categorized into three levels of complexity. We adhere to the aforementioned definition by simulating terrains with varying topographical complexities. Topographical complexity is defined by ground roughness and slope. The ground in the easy plot is more flat, whereas the ground in the difficult plot is considerably rougher (see Figure \ref{fig:data_construction}). Consequently, a terrain is generated for each simulated plot and presented as a GeoTIFF file with a grid resolution of 0.2 meters.

\subsubsection{Understory Generation}
The understory, comprising grasses and shrubs, is a vital component of forest ecosystems. Nevertheless, the creation of a mesh model of the understory is a particularly challenging task, due to the prevalence of intricate micro-structures within this area. In this study, we employ a straightforward approach to preserve the understory of forest plots in the EuroSDR benchmark projects. Firstly, the graph pathing algorithm \citep{WANG202167} is employed to segment the understory points from the provided benchmark plot data. The segmented understory points, augmented with a variety of strategies, including random scaling and rotation, are directly input into Helios++, which is capable of accepting such diverse inputs. It should be noted that the aforementioned understory points are cropped and placed on the generated terrain in accordance with the extent and topography of the latter.

\subsubsection{Plot Creation}
\label{plotcreation}
Once the tree model database, terrain, and understory have been created, the subsequent step is to assemble them into a comprehensive plot. The crucial point is the decision to place individual trees on the terrain. It is of particular importance that the workflow of our digital cousins resembles the geometric properties observed in real forest plots. Consequently, particular attention is paid to the resemblance of stem density and tree height distributions. 

Specifically, the initial step is to define the number of trees to be placed in a virtual plot. This is achieved through random sampling of a number that falls within the range of the average number of trees observed in the real plot, plus or minus one standard deviation. For the variable of tree height, a uniform distribution is sampled according to the mean and standard deviation of the real plot. Subsequently, a tree model from the database is randomly selected with a closed height value. The selected tree model is then randomly scaled and rotated. Its location is randomly set with the terrain extent. When placing a new tree, a maximum of 5\% canopy overlapping ratio is permitted, same as \cite{guzman2020relationship}. This procedure is continued until the desired number of trees have been placed. 

However, although this strategy works for easy and medium complexity plots, it is unsuitable for difficult plots as these plots tend to be multi-layered rather than single-layered. In order to account for such copy complexity, we have also developed a strategy for creating multi-layered difficult plots. The target trees in the difficult plots are split into two groups, while together they still follow the stem density and height distribution of the real-world difficult plots. Here, only half of the target trees that are tall are placed using the same procedure as for the easy and medium plots. The other half, small trees, are then planted gradually under the large trees to create a multi-layered forest. Throughout the production process, we continue to constrain the canopy overlap ratio, rather than simply placing them all at once. This ensures that the virtual plots created retain the geomorphic characteristics and ecological rules of the real plots. A comparison of plot statistics between generated and real plots can be found in section \ref{sec:Statistics}. A total of 1,000 plots with 20 x 20 m size were created, comprising 334 for the easy plot category, 333 for the medium and difficult categories, respectively.

\subsection{Multi-platform LiDAR Point Cloud Simulation}
Four commonly used LiDAR platforms for forest applications including TLS, MLS, ULS and ALS  are simulated in this study. SLS is not considered here as our focus is on the fine-grained structure of forest at the individual tree of plot scale, while SLS is targeted for ultra-large scale studies.

\subsubsection{TLS}
\begin{figure}
	\centering
	\includegraphics[width=.8\textwidth]{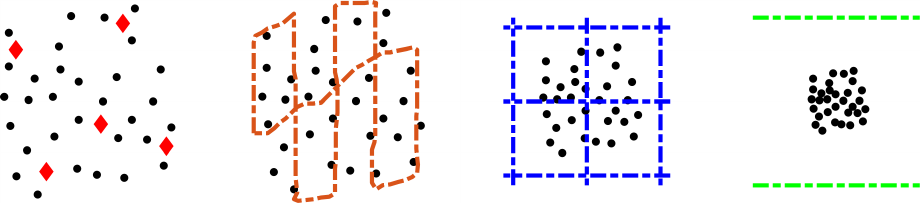}
	\caption{The scan setups for each platform, from left to right: TLS, MLS, ULS and ALS. The black dots indicate the individual tree locations within a plot, while the red diamonds in TLS represent the five scanning locations. The dot lines in MLS depict the automatically planned moving trajectory, and the blue lines represent the tic-tac-toe trajectories for ULS. The green lines illustrate the flight trajectories for ALS.}
	\label{fig:scan_setups}
\end{figure}

The simulated TLS scanner is the Riegl VZ-400 (Riegl Laser
Measurement Systems GmbH, Austria), which is a commonly recommended choice for the study of fine-grained forest structures. The physical properties of this scanner are accessible in Helios++ by default. In this instance, both the horizontal and vertical scanning resolutions were set at 0.04\degree. The mounting platform is a static tripod, with the scanner positioned at an elevation of 1.5 meters above ground level. Five scanning locations are simulated, with one in the plot center and four in the corners. This setup is the most commonly employed scanning setup for plot-scale TLS surveys. The objective is to optimize the scanning locations so that they do not conflict with the tree positions. A simple grid search method can be employed to achieve this goal. An example of optimized TLS scan locations is shown in Figure \ref{fig:scan_setups}. 
\subsubsection{MLS}
MLS systems, such as backpack or handheld LiDAR systems, are increasingly utilized in forest studies due to their high mobility and capacity to acquire point clouds in environments where GPS is unavailable. These systems are based on the technology of simultaneous localization and mapping (SLAM). In this study, we simulate the Velodyne VLP-16 scanner (Velodyne Lidar, Inc, USA), which is a popular choice among commercial MLS devices. In contrast to other survey-grade scanners, the VLP-16 is a multi-channel scanner that is primarily used for autonomous driving applications. Additionally, configurations of such multi-channel scanners, including the VLP-16, are available for review in the comprehensive scanner database of Helios++. The scanner is affixed to a mobile platform situated at an elevation of 1.8 meters above ground, thereby approximating the position of a person carrying the scanner. The simulated working speed is 1.3 meters per second.
A fundamental aspect of MLS simulation is the design of a route that closely resembles the walking path typically observed in real-world surveys. An automatic path planning method was developed, whereby several turning points are first defined along the plot borders. Subsequently, the spaces occupied by tree stems are masked with a buffer distance of 0.5 meters around them. This is done to circumvent potential conflicts with tree stems when traversing the plot. Subsequently, the shortest path algorithm is employed to identify the optimal route between two successive turning points. Furthermore, a loop closure was implemented to reflect the actual survey plans in SLAM-based MLS systems. Figure \ref{fig:scan_setups} illustrates an example of an automatically planned MLS trajectory.

\subsubsection{ULS}
Unmanned aerial vehicles (UAVs) equipped with ULS systems are utilized for the purpose of scanning forests. It is a promising platform with the potential to efficiently acquire dense point clouds over a large region. The Riegl UAV-1 (Riegl Laser
Measurement Systems GmbH, Austria) is a frequently selected option for forest applications, and its comprehensive technical specifications are also accessible in Helios++ by default. As ULS operate above the forest, the simulation of ULS is straightforward: the platform is fixed at a height of 50 meters above ground, and a tic-tac-toe trajectory pattern with a distance of 15 meters between neighboring trajectories is configured. In this instance, the speed of the ULS system in motion is 5 meters per second.

\subsubsection{ALS}
The ALS system is employed on manned aircraft for the purpose of scanning forest areas across a vast expanse. In ALS, the aircraft is flown at a significantly higher altitude, resulting in a considerably sparser point cloud. In this study, we conduct a simulation of the Riegl LMS-Q780 scanner (Riegl Laser
Measurement Systems GmbH, Austria), which is mounted on an aircraft flying at an altitude of 800 meters and a speed of 45 meters per second. As illustrated in Figure \ref{fig:scan_setups}, two parallel flight trajectories, with a distance of 60 meters between them, were simulated.

\subsubsection{Point Cloud Consolidations}
\begin{figure}
	\centering
	\includegraphics[width=\textwidth]{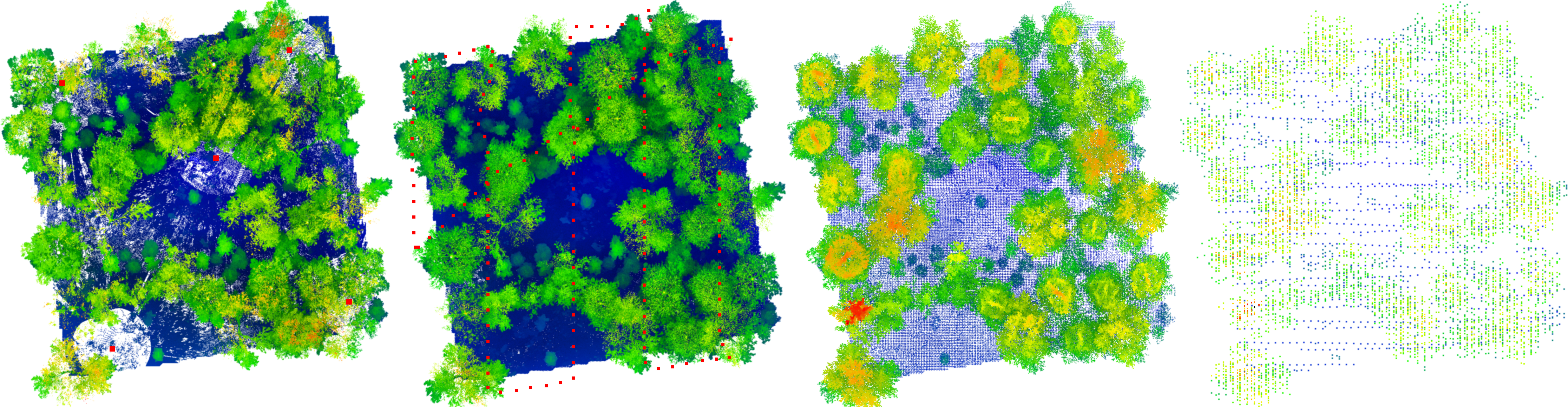}
	\caption{Top review of simulated multi-platform point clouds. From left to right: TLS, MLS, ULS and ALS. The red dots in TLS and MLS represent the scan locations and moving trajectories, respectively. }
	\label{fig:top_review}
\end{figure}
The simulation framework yields multi-platform point cloud data for 1,000 forest plots (see Figure \ref{fig:top_review} for an example), which are subsequently consolidated and organized. For each plot, the consolidated point clouds from all platforms contain ground-truth labels of both semantic (e.g., wood, leaf, understory, and terrain) and instance (e.g., single tree) categories. Additionally, a label indicating the scanning configuration for each platform is also available. In the case of TLS, this refers to the scanning location, whereas in the case of MLS, this refers to a specific way point along the walking trajectory. In the case of ULS and ALS, the aforementioned label serves to indicate the specific flight trajectory from which the data originates. All point clouds were stored in the LAZ format, a compressed version of LAS format that optimizes storage efficiency, with the ground-truth labels stored as additional fields in the LAZ files. The large-scale synthetic point cloud dataset of forest scenes, which has been constructed, is named Boreal3D.

\section{Statistics of Boreal3D}
\label{sec:Statistics}

\subsection{Point Distributions}
\label{sec:Point_distributions}

\begin{figure}
	\centering
	\subfigure[]{
		\includegraphics[width=0.3\textwidth]{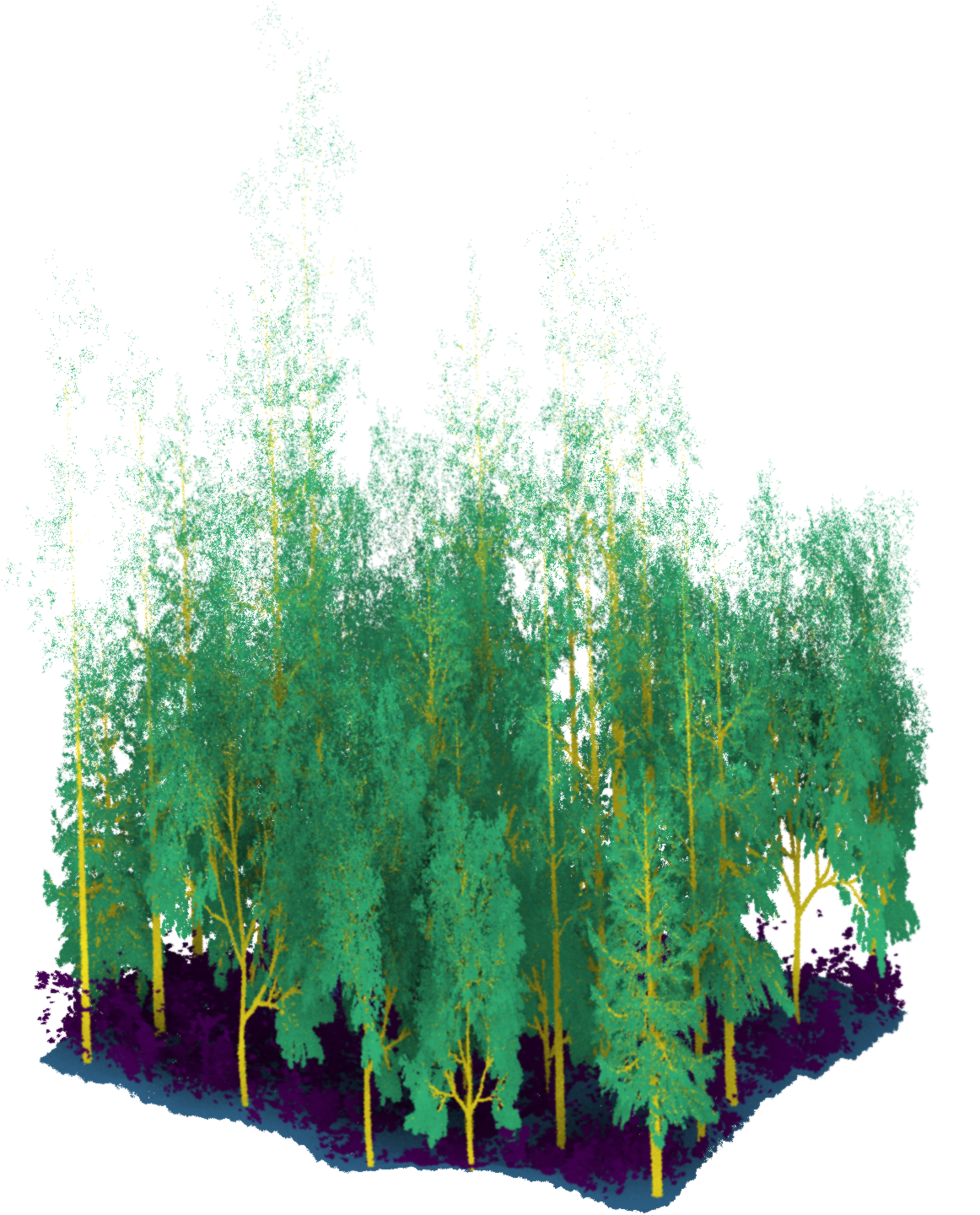}
	}
	\subfigure[]{
		\includegraphics[width=0.3\textwidth]{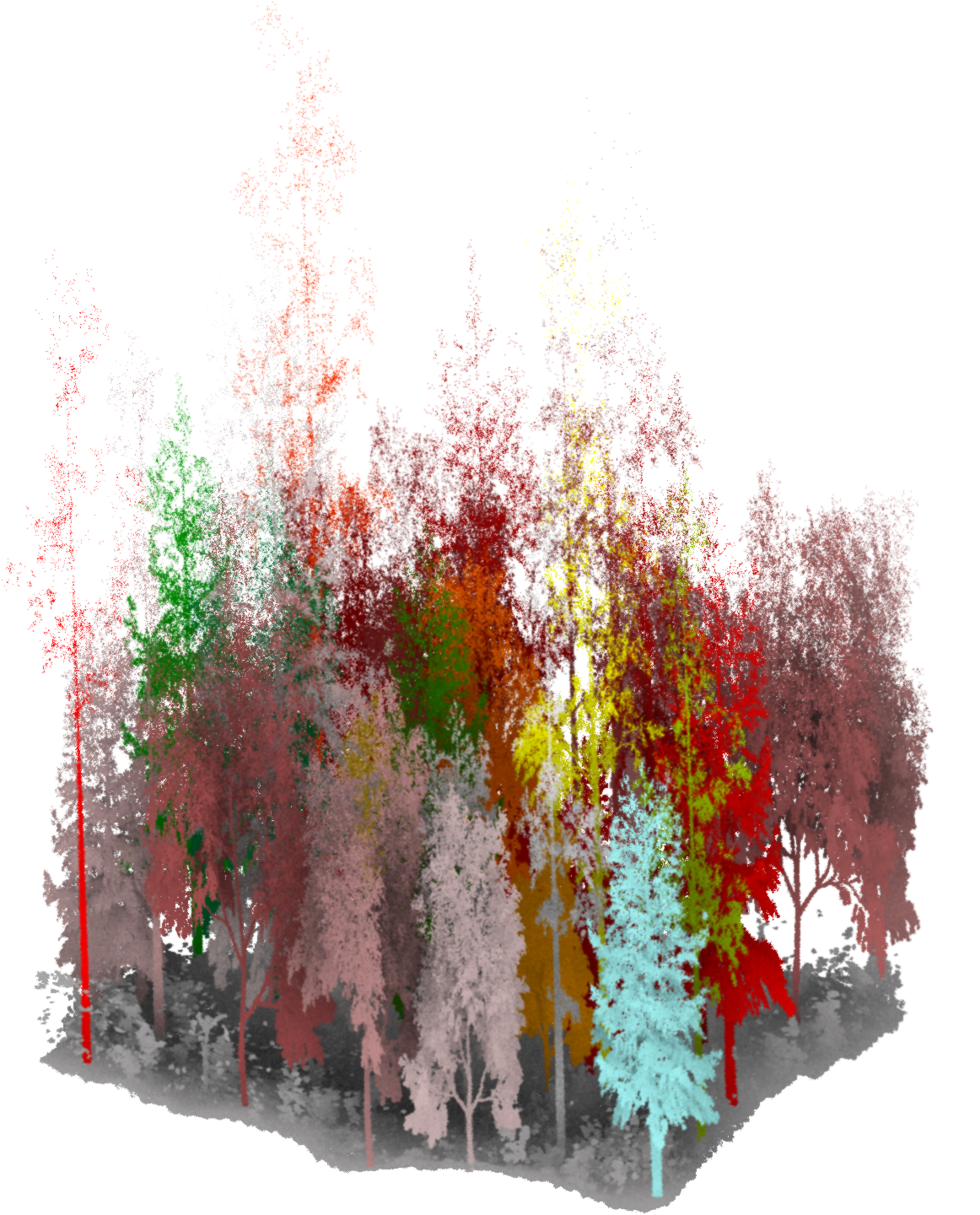}
	}
	\caption{Visualization of point cloud in a plot with semantic and instance annotations. (a) represents semantic annotations, and (b) represents instance annotations. Labels are colored randomly.}
	\label{fig:VisualizationSemanticInstance}
\end{figure}

\begin{figure}
	\centering
	\includegraphics[width=0.4\textwidth]{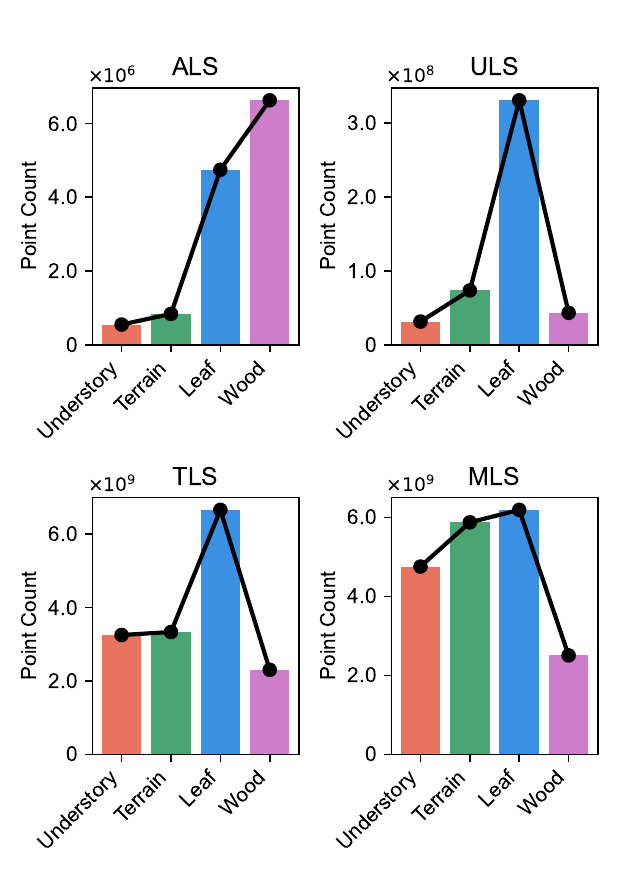}
	\caption{The number of points for each semantic category in different platforms}
	\label{fig:point_counts_with_scientific}
\end{figure}

Boreal3D provides both semantic and instance labels for each point. The semantic labels include four classes: terrain, understory, leaf, and wood, as shown in Figure \ref{fig:VisualizationSemanticInstance} (a). Table \ref{IDSemanticLabel} and Figure \ref{fig:point_counts_with_scientific} illustrate the distribution of points in each semantic class across different platforms.  It can be clearly seen from Figure \ref{fig:point_counts_with_scientific} that, affected by the acquisition characteristics of different platforms, the proportion of different semantics in each platform is quite different. Overall, the number of point clouds across platforms follows a specific order, with ALS having the fewest, followed by ULS, then TLS, and finally MLS having the highest count. Secondly, the number of points in the understory class is the least in ALS and ULS, while the number of understory in TLS and MLS has increased significantly. In addition, even for acquisition methods with similar characteristics (for example, ALS and ULS, TLS and MLS), the proportion of the number of points in each semantic class is very different. This illustrates the diverse point cloud properties across different LiDAR platforms.

\begin{table}[t]
	\centering
	\resizebox{0.5\textwidth}{!}{
	\begin{tabular}{ccccc}
		\toprule
		 & Understory & Terrain & Leaf & Wood \\
		\midrule
		ALS & 548265 & 838594 & 4747851 & 6632251 \\
		ULS & 31418105 & 73682290 & 331199714 & 43297359 \\
		TLS & 3253558995 & 3336100208 & 6663985883 & 2303878571 \\
		MLS & 4758072765 & 5877772240 & 6193336270 & 2506508913 \\
		\bottomrule
	\end{tabular}
	}
	\caption{The distribution of points across different platforms for each semantic class.}
    \label{IDSemanticLabel}
\end{table}

For instance labels, all points in the non-tree part of the plot (corresponding to "understory" and "terrain" in semantics) are assigned a value of 0, and other points are assigned values of 1, 2, 3, etc. according to the tree they belong to (Figure \ref{fig:VisualizationSemanticInstance} (b)). In more detail, in addition to these conventional instance labels, Boreal3D also provides the location coordinates and tree species labels for each tree. Table \ref{tab:differentTreeSpecies} details the instance counts for the three tree species included in the dataset. It can be seen that the distribution of tree species is relatively balanced across the entire dataset. Additionally, this dataset provides viewpoint information for every point, offering robust support for tasks such as point cloud registration and other 3D understanding challenges. 

\begin{table}[t]
	\centering
	\resizebox{0.3\linewidth}{!}{
	\begin{tabular}{cccc}
	\toprule
	Species & Pine & Birch & Spruce \\
	\midrule
	Tree count  & 14435 & 16483 & 17485 \\
	\bottomrule
	\end{tabular}
	}
        \caption{The number of trees for three different species.}
        \label{tab:differentTreeSpecies}
\end{table}
\subsection{Tree Attributes Statistics}
\begin{table}[t]
	\centering
	\resizebox{0.65\linewidth}{!}{
	\begin{tabular}{ccccccc}
	\toprule
	 & \multicolumn{2}{c}{Stem density ($stem/ha$)} & \multicolumn{2}{c}{DBH ($cm$)} & \multicolumn{2}{c}{Tree height ($m$)} \\
      Complexity categories & EuroSDR & Boreal3D & EuroSDR & Boreal3D & EuroSDR & Boreal3D \\
	\midrule
	Easy        & 592$\pm$ 189  & 613$\pm$ 114   & 20.7$\pm$ 8.5  & 16.2$\pm$ 9.1 & 18.4$\pm$ 6.4 & 20.7$\pm$ 6.2   \\
    Medium      & 968$\pm$ 370  & 1005$\pm$ 210  & 17.2$\pm$ 10.7 & 13.9$\pm$ 7.8 & 16.2$\pm$ 7.3 & 21.0$\pm$ 6.1     \\
    Difficult   & 2021$\pm$ 553 & 2013$\pm$ 314  & 12.3$\pm$ 7.2  & 9.2$\pm$ 7.1 & 13.2$\pm$ 5.9  & 16.3$\pm$ 9.1    \\
	\bottomrule
	\end{tabular}
	}
        \caption{Comparison of Boreal3D and EuroSDR benchmark dataset on key plot attributes.}
        \label{tab:boreal3dvsEuroSDR}
\end{table}

\begin{figure}
	\centering
	\includegraphics[width=0.45\linewidth]{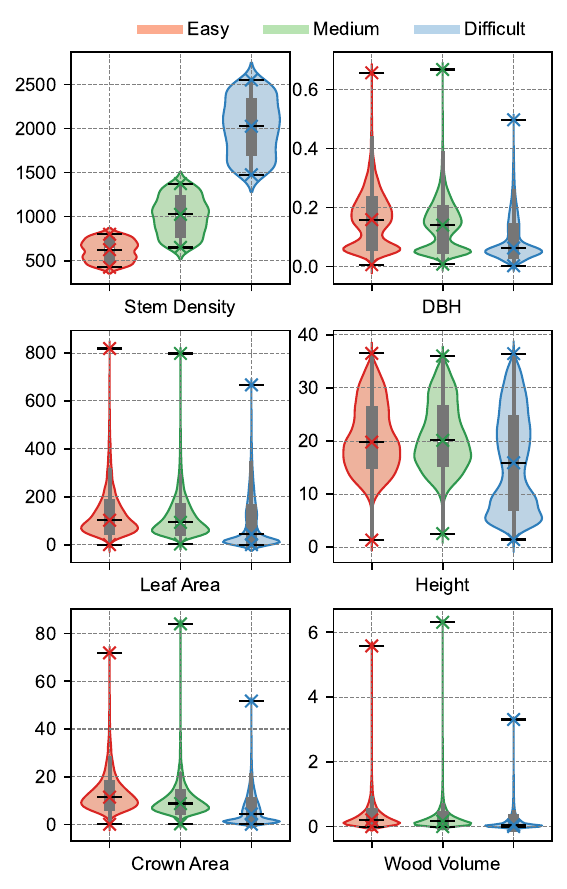}
	\caption{Violin plots of six main attributes for forest plots with different complexities. Units are: Tree Density ($stem/ha$), DBH ($m$), Leaf Area ($m^2$), Height ($m$), Crown Area ($m^2$), Wood Volume ($m^3$).}
	\label{fig:AttributeStatisticsChart}
\end{figure}

\begin{figure}
	\centering
	\includegraphics[width=0.8\linewidth]{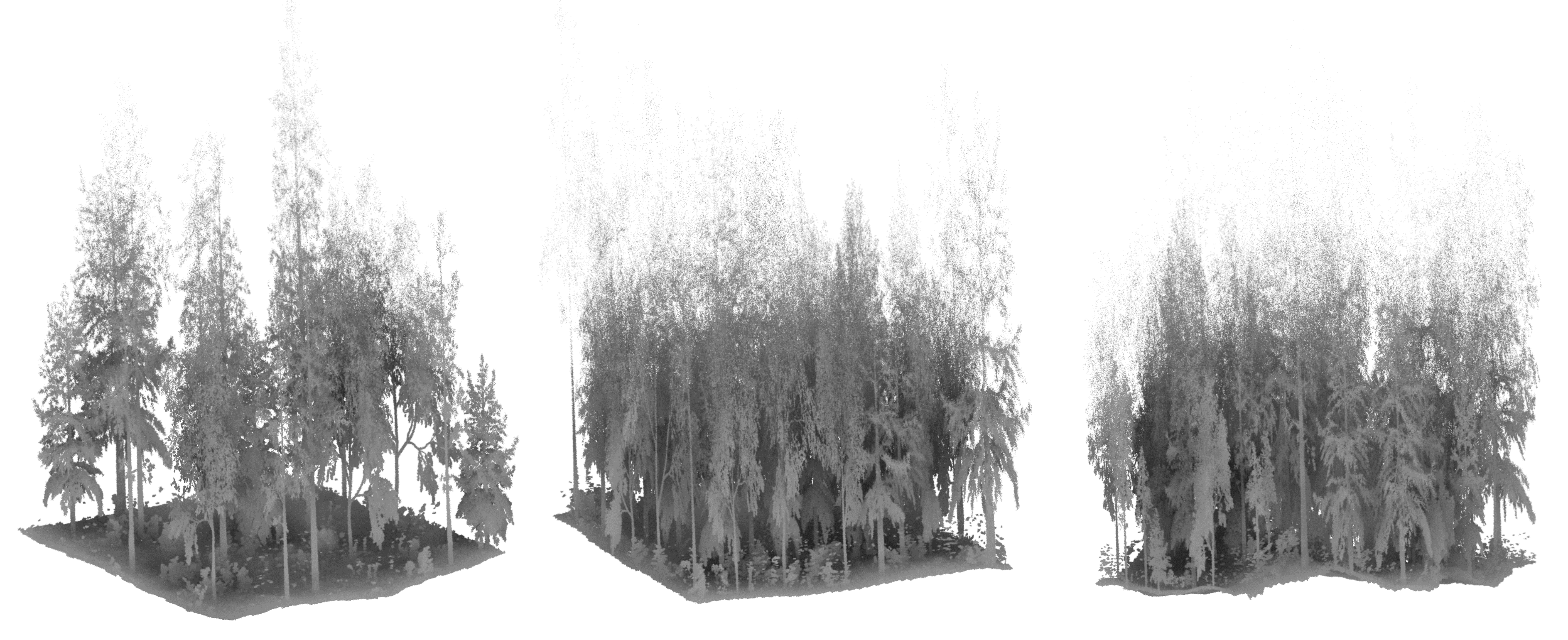}
	\caption{Visualization example of TLS point clouds for plots with different complexities. From left to right, Easy plots, Medium plots, and Difficult plots.}
	\label{fig:treeTreeDensity}
\end{figure}

In addition to point-wise labels and information, Boreal3D also provides key structural parameters for each tree, including height, DBH, crown width, leaf area, and volume. These parameters can be used to explore research on the estimation of structural metrics at tree and forest level. Figure \ref{fig:AttributeStatisticsChart} provides a comparative visualization of six different structural parameters from the perspective of different plot complexities. The three plot categories exhibit differences in DBH, leaf area, tree height, crown area, and wood volume. Specifically, it can be clearly seen from Figure \ref{fig:AttributeStatisticsChart} that the density of trees in the three types of forests gradually increases from easy to medium to difficult. The sparse easy plot has the most trees with an DBH greater than 0.1 m, while the dense difficult plot has smaller DBH values. From the distribution of tree height, it can be seen that the trees in the easy and medium plots are mainly concentrated between 10-40 m, while the difficult plots additionally consider trees less than 10 m to construct complex multi-layer plots. Subject to the same influence, leaf area, crown area and wood volume have similar distributions. Table \ref{tab:boreal3dvsEuroSDR} presents a comparative analysis between the Boreal3D and EuroSDR benchmark dataset \citep{liang2018international}, from which we generated the digital cousins, focusing on three critical attributes: stem density, DBH, and tree height. The analysis reveals that the Boreal3D dataset, while exhibiting measurable distinctions, effectively maintains key attributes that closely approximate real-world forest conditions. Figure \ref{fig:treeTreeDensity} illustrates visualizations of simulated TLS point cloud of three forest plots with varying levels of complexity. 

\subsection{Data Partitions}

To facilitate future deep learning based research on fine-grained 3D forest structure understanding tasks such as semantic segmentation and instance segmentation, the dataset is split into training, validation, and test sets in a 6:2:2 ratio. The specific split details are shown in Table \ref{tab:trainvaltest}. In the specific partition process, we mainly consider the number of plots with three different complexities. According to the number of plots, the training and validation sets have 603 and 199 plots respectively, and the test set contains 198 plots. The number of plots of each complexities in each set is basically uniform. It can also be seen from Table \ref{tab:trainvaltest} that the number of trees in the divided plot sets basically meets the requirements.

Overall, Boreal3D comprises a total of 1,000 forest plots. Each plot includes point clouds collected from four different platforms, with each point annotated with semantic, instance, and viewpoint labels. Additionally, each tree is accompanied by ground-truth data on several key structural attributes.
\begin{table}
	\centering
	\resizebox{0.45\textwidth}{!}{
	\begin{tabular}{l|ccc|ccc|ccc}
	\toprule
	& \multicolumn{3}{c|}{Train ($\approx$60\%)} & \multicolumn{3}{c|}{Val ($\approx$20\%)} & \multicolumn{3}{c}{Test ($\approx$20\%)} \\
	\midrule
	& Plots & Trees & & Plots & Trees & & Plots & Trees & \\
	\midrule
	Easy & 201 & 4949 & & 67 & 1650 & & 66 & 1601 & \\
	Medium & 201 & 8095 & & 66 & 2590 & & 66 & 2701 & \\
	Difficult & 201 & 16129 & & 66 & 5401 & & 66 & 5287 & \\
	Total & 603 & 29173 & & 199 & 9641 & & 198 & 9589 & \\
	\bottomrule
	\end{tabular}
	}
        \caption{Statistics of the training, validation, and test data splits.}
        \label{tab:trainvaltest}
\end{table}

\section{Experiment}
\label{sec:Experiment}
The primary objective of this study is to enhance our capacity for understanding fine-grained 3D forest structures through the application of digital cousins and Sim2Real techniques. A significant focus of this research lies in the potential of simulated point clouds to facilitate advancements in and enhance the efficacy of state-of-the-art deep learning methodologies for forest structure analysis. To validate this conceptual framework, a series of meticulously designed experiments were conducted, the results of which are presented and analyzed in this work.

\subsection{Experiment Setup}
\subsubsection{Applicability of Boreal3D for Real-World Forest Point Cloud Analysis}
\label{transfer}

Initially, we conduct a comprehensive investigation into the transferability of Boreal3D to real-world datasets, with a specific focus on the For-instance dataset \citep{puliti2023instance}. The For-instance dataset comprises ULS data collected across five distinct geographical regions. This dataset was selected as our benchmark for two primary reasons: Firstly, it provides manually annotated labels for both semantic and instance segmentation, enabling simultaneous validation of these two fundamental tasks. Secondly, For-instance has been extensively used in recent studies for developing deep learning models in forest applications \citep{WIELGOSZ2024114367, XIANG2024114078}, thereby establishing its reliability and quality within the research community. 

The benchmark evaluation focuses on two fundamental tasks: semantic and instance segmentation. Given that the primary objective of this study is not to develop novel deep learning architectures but rather to assess dataset transferability, we employ well-established, state-of-the-art models that represent current methodological advancements. For semantic segmentation, we utilize RandLA-Net \citep{hu2020randla}, known for its efficient large-scale point cloud processing, and Point Transformer v3 (PTv3) \citep{wu2024point}, which demonstrates advanced transformer-based capabilities. Regarding instance segmentation, our selection includes TreeLearn \citep{henrich2024treelearn}, specifically designed for forest structure analysis, and OneFormer3D \citep{kolodiazhnyi2024oneformer3d}, a transformer-based approach that has shown superior performance in 3D instance segmentation tasks.

Specifically, we devise and implement five distinct options to evaluate the applicability of Boreal3D in both semantic and instance segmentation tasks utilizing deep learning frameworks:
\begin{itemize}
  \item \textit{Real to Real Scenario:} In this configuration, both training and testing are exclusively performed on the For-instance dataset. This approach establishes a baseline performance metric for models trained on annotated real-world point clouds, serving as a critical reference point for evaluating the effectiveness of simulated data in subsequent experiments.

  \item \textit{Sync to Real Scenario:} In this configuration, the synthetic Boreal3D dataset (only ULS part) is utilized exclusively for model training, while evaluation is performed on the real-world For-instance dataset. This approach directly assesses the transferability of knowledge learned from synthetic data to real-world applications, providing critical insights into the efficacy of Sim2Real methodologies in forest structure analysis.
  
  \item \textit{Sync-Real Mixed Scenario:} In this configuration, both the synthetic Boreal3D dataset and the real-world For-instance dataset are combined for model training, while evaluation is conducted exclusively on the For-instance dataset. This hybrid approach aims to investigate the potential benefits of augmenting real-world data with synthetic samples, thereby exploring whether the integration of diverse data sources can enhance model performance and generalization capabilities in forest structure analysis.
  
  \item \textit{Fine-tuning Sync with Real Scenario:} In this configuration, models are first pre-trained on the synthetic Boreal3D dataset and subsequently fine-tuned using the real-world For-instance dataset, with evaluation performed exclusively on the For-instance dataset. This approach examines the effectiveness of leveraging synthetic data as a foundation for pre-training, followed by domain-specific refinement using real-world data, thereby exploring its potential to improve model performance and adaptability in forest structure analysis.
  
  \item \textit{Fine-Tuning with Varied Data Sizes Scenario:} In addition, following the \textit{Fine-tuning Sync with Real Scenario}, we also examine the impact of real-world data quantity on model performance.
\end{itemize}

The metrics employed for semantic and instance segmentation align with established standards in the literature \citep{henrich2024treelearn,WIELGOSZ2024114367,  XIANG2024114078}. For semantic segmentation, we utilize Overall Accuracy (OA), mean Accuracy (mACC), and mean Intersection over Union (mIoU). For instance segmentation, the evaluation includes Completeness, Omission Error, Commission Error, and F1-Score. These metrics are formally defined as follows:

\begin{equation}
    Precision = \frac{TP}{TP + FP},
\end{equation}
\begin{equation}
    Recall = \frac{TP}{TP + FN},
\end{equation}
\begin{equation}
    Accuracy = \frac{TP + TN}{TP + TN + FP + FN},
\end{equation}
\begin{equation}
    F1-Score = 2\times \frac{Precision\times Recall}{Precision + Recall},
\end{equation}
where TP represents true positives, FP denotes false positives, FN stands for false negatives, and TN indicates true negatives.

Let $C$ be the confusion matrix with n classes. The OA is defined as:
\begin{equation}
    OA = \frac{\sum_{i=1}^n{C_{ii}}}{\sum_{i=1}^n{\sum_{j=1}^n{C_{ij}}}},
\end{equation}
\begin{equation}
    mACC = \frac{1}{n}\sum_{i=1}^n{Accuracy_i},
\end{equation}
where n refers to the total number of categories.
\begin{equation}
    IoU = \frac{GT \cap Pred }{GT \cup Pred},
\end{equation}
\begin{equation}
    mIoU = \frac{1}{n}\sum_{i=1}^n{IoU_i},
\end{equation}
where n refers to the total number of categories.
\begin{equation}
    Completeness = \frac{TP}{TP + FP} = Recall,
\end{equation}
\begin{equation}
    Omission\ error = 1 - \frac{TP}{TP + FP} = 1 - Recall,
\end{equation}
\begin{equation}
    Commission\ error = 1 - \frac{FP}{TP + FP} = 1 - Precision,
\end{equation}
\begin{equation}
    Kappa = \frac{Accuracy - P_c}{1 - P_c},
\end{equation}
where:
\begin{equation}
    P_c = \frac{(TN+FN) * (TN + FP) + (FN + TP)*(FP + TP) }{(TP + TN + FP + FN)^2}.
\end{equation}

\noindent It should be noted that the IoU threshold for matching the predicted mask and the ground-truth mask in the instance segmentation is 0.5. Detailed explanations of above used metrics for forest point cloud segmentation can be found in \cite{XIANG2024114078} and \cite{henrich2024treelearn}.

\subsubsection{Applicability of Boreal3D for Cross-platform Forest Point Cloud Analysis}
The comprehensive experiments conducted on the transferability of Boreal3D to real-world datasets provide critical insights into whether Boreal3D, and more broadly, Sim2Real techniques, can enhance the effectiveness of real-world forest structure analysis. These findings will not only evaluate the specific utility of Boreal3D but also contribute to the broader understanding of synthetic-to-real transfer learning methodologies in forest-related applications. However, these experiments are exclusively conducted on the ULS For-instance dataset. While extending such evaluations to other platforms would significantly enhance the assessment of cross-platform applicability, the generation of well-annotated real-world datasets remains a considerable challenge. To further investigate the utility of Boreal3D across diverse point cloud sources and their associated tasks, we have expanded our experimental framework to include ALS, MLS and TLS platforms.

Specifically, we curated open-source real-world datasets corresponding to each of these platforms and designed targeted experiments to evaluate platform-specific tasks. For each, point clouds from the corresponding platform in Boreal3D are used for model pre-training, followed by fine-tuning using real-world datasets. To ensure rigorous and unbiased evaluation, we maintain strict separation of the real-world test sets, preventing any exposure during the training or fine-tuning phases, thereby guaranteeing a fair and reliable comparison across all experimental conditions. Metrics used for these experiments are the same as in section \ref{transfer}.

\begin{itemize}
  \item \textit{ALS:} ALS is widely employed for large-scale forest applications due to its extensive coverage capabilities. Here, we focus on ground filtering as the primary task, utilizing the OpenGF dataset \citep{qin2021opengf}, a large-scale ALS point cloud dataset specifically designed for ground filtering evaluation. The experimental results are compared with those reported in \citet{rs14225798}, in which a deep learning method designed specifically for forest ground filtering is proposed.

  \item \textit{MLS:} As for MLS, we employ the Point2Tree (P2T) dataset \citep{rs15153737} as the real-world benchmark. Collected in 2022, this dataset comprises 16 MLS point cloud samples of circular plots, providing a robust foundation for evaluating instance segmentation tasks. The experimental results are compared with those reported in \cite{rs14236116}.
  
  \item \textit{TLS:} Regarding TLS, we adopt the dataset from \cite{tls-wood-leaf} as the real-world benchmark for our experiments. We focus on leaf-wood separation, a critical and long-standing challenge in forest point cloud analysis and a specialized case of semantic segmentation. This dataset comprises TLS point clouds of three distinct tree species: White birch (\textit{Betula platyphylla}), Dahurian larch (\textit{Larix gmelinii}), and Chinese scholar tree (\textit{Sophora japonica}). Results are compared with those from the original study in \cite{tls-wood-leaf}.

\end{itemize}
\subsection{Results}

\begin{table}[t]
    \centering
    \resizebox{0.6\linewidth}{!}{
    \begin{tabular}{ccccccc}
    \toprule
        & \multicolumn{3}{c}{RandlA-Net \citep{hu2020randla}} & \multicolumn{3}{c}{PTv3 \citep{wu2024point}} \\
        & OA (\%) & mACC (\%) & mIoU (\%) & OA (\%) & mACC (\%) & mIoU (\%) \\
    \midrule
    Real only & 90.45 & 85.12 & 75.71 & 92.69 & 85.80 & 78.59 \\
    Sync only  & 73.16\color{green}{\color{green}{$\downarrow$}} & 57.48\color{green}{$\downarrow$} & 39.97\color{green}{$\downarrow$} & 84.41\color{green}{$\downarrow$} & 66.02\color{green}{$\downarrow$} & 55.28\color{green}{$\downarrow$} \\
    Sync + Real & 89.93\color{green}{$\downarrow$} & 76.78\color{green}{$\downarrow$} & 68.69\color{green}{$\downarrow$} & 92.86\color{red}{$\uparrow$} & 91.22\color{red}{$\uparrow$} & 81.05\color{red}{$\uparrow$} \\
    Sync + Fine-tuning & 91.65\color{red}{$\uparrow$} & 87.45\color{red}{$\uparrow$} & 76.96\color{red}{$\uparrow$} & 93.61\color{red}{$\uparrow$} & 91.30\color{red}{$\uparrow$} & 82.40\color{red}{$\uparrow$} \\
    \bottomrule
    \end{tabular}
    }
    \caption{
        Semantic segmentation results under different training scenarios.
    }
    \label{tab:semantic_results}
\end{table}

\begin{figure}
	\centering
	\includegraphics[width=\linewidth]{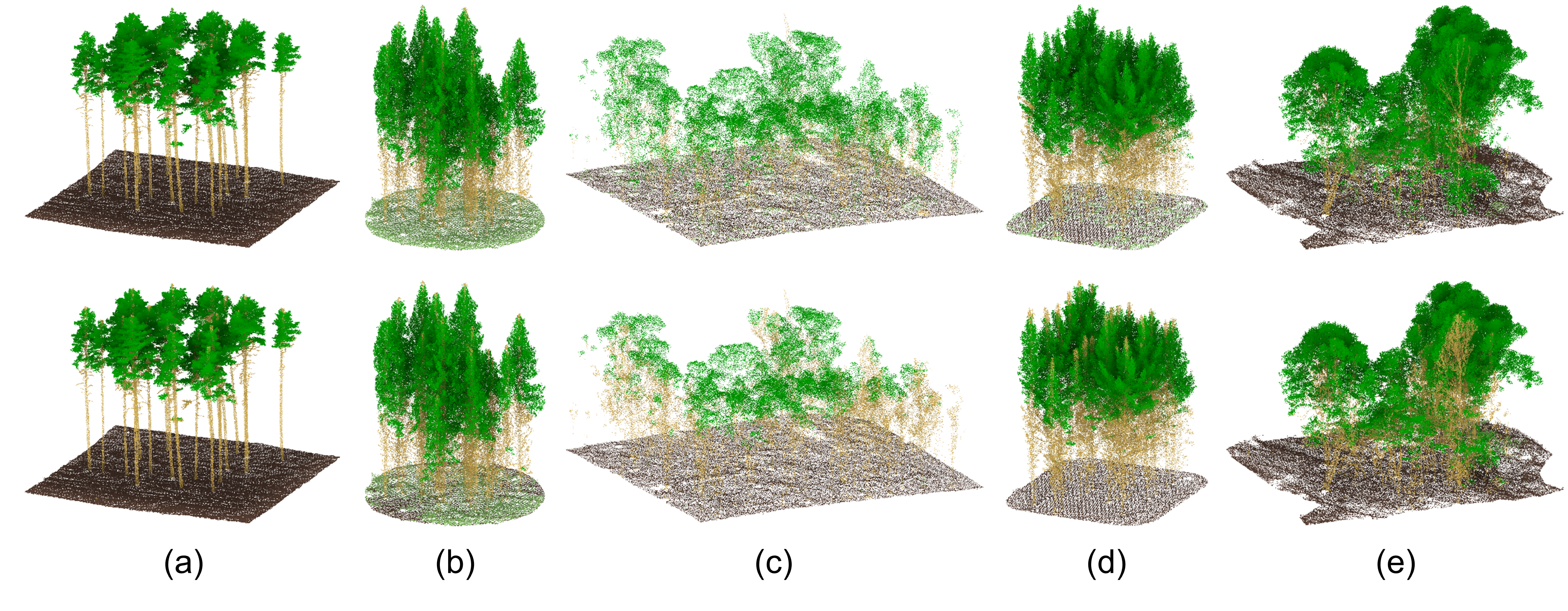}
	\caption{Visualization of semantic segmentation results. The top row represents the ground truth, while the bottom row displays the predictions generated by the fine-tuned model.}
    \label{fig:seman}
\end{figure}

\subsubsection{Boreal3D for Real-World Forest Point Cloud Analysis}
Table \ref{tab:semantic_results} presents the performance comparison across different scenarios for the semantic segmentation task. When exclusively using Boreal3D as the training data, the models exhibit a significant performance decline, highlighting the inherent domain gap between synthetic and real-world data. However, when Boreal3D is combined with real-world datasets, model performance improves substantially. In this mixed training scenario, PTv3 achieves an OA of 92.86\% and a mIoU of 81.05\%, surpassing the results obtained from models trained solely on real-world datasets. Notably, fine-tuning models pre-trained on Boreal3D yields exceptional results: RandLA-Net attains an OA of 91.65\%, while PTv3 achieves an OA of 93.61\%, both significantly outperforming models trained exclusively on real-world data. Particularly impressive is PTv3's improvement of 3.81\% in mIoU, demonstrating the effectiveness of synthetic pre-training followed by real-world fine-tuning. Figure \ref{fig:seman} shows the segmentation result of PTv3 prediction after fine-tuning. For more detailed results on the subsets of the For-instance dataset, please refer to the supplementary materials.

\begin{table}[t]
\resizebox{\linewidth}{!}{
    \centering
    \begin{tabular}{ccccccccc}
    \toprule
        & \multicolumn{4}{c}{TreeLearn \citep{henrich2024treelearn}} & \multicolumn{4}{c}{Oneformer3D \citep{kolodiazhnyi2024oneformer3d}} \\
        & Completeness (\%)  & Omission error (\%) & Commission error (\%) & F1-score (\%) &  Completeness (\%)  & Omission error (\%) & Commission error (\%) & F1-score (\%)  \\
    \midrule
    Real only & 83.2 & 16.9 & 24.3 & 78.6  & 59.0 & 41.1 & 25.0 & 66.0 \\
    Sync only & - & - & - & -    & 48.5\color{green}{\color{green}{$\downarrow$}}  & 51.5\color{green}{\color{green}{$\downarrow$}}  & 42.8\color{green}{\color{green}{$\downarrow$}}  & 52.5\color{green}{\color{green}{$\downarrow$}}  \\
    Sync + Real             & 19.4\color{green}{\color{green}{$\downarrow$}}  & 80.6\color{green}{\color{green}{$\downarrow$}}  & 84.0\color{green}{\color{green}{$\downarrow$}}  & 17.3\color{green}{\color{green}{$\downarrow$}}  & 67.3\color{red}{$\uparrow$} & 32.8\color{red}{$\uparrow$} & 19.4\color{red}{$\uparrow$} & 73.3\color{red}{$\uparrow$} \\
    Sync + Fine-tuning        & 81.4\color{green}{\color{green}{$\downarrow$}}  & 18.6\color{green}{\color{green}{$\downarrow$}}  & 15.9\color{red}{$\uparrow$} & 82.3\color{red}{$\uparrow$}  & 70.3\color{red}{$\uparrow$} & 29.6\color{red}{$\uparrow$} & 25.8\color{green}{\color{green}{$\downarrow$}}  & 72.2\color{red}{$\uparrow$} \\
    \bottomrule
    \end{tabular}
}
    \caption{
    Instance segmentation results under different training scenarios.
    }
    \label{tab:instance_results}
\end{table}

\begin{figure}
	\centering
	\includegraphics[width=0.8\linewidth]{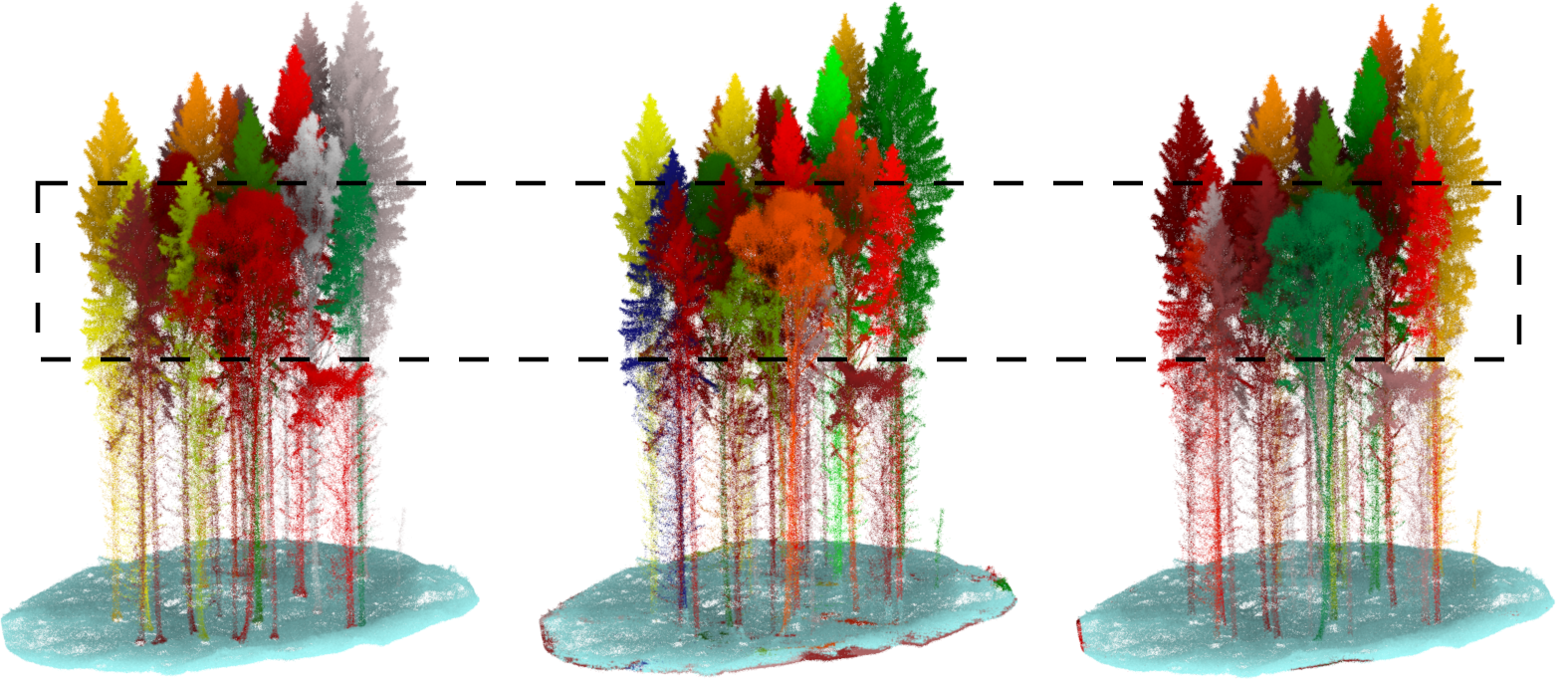}
	\caption{Visualization of instance segmentation results. From left to right: (1) the ground-truth annotations, (2) the segmentation results obtained from a model trained exclusively on real-world data, and (3) the improved results achieved through fine-tuning a pre-trained model. The black rectangle is superimposed on the visualization to highlight specific regions where the fine-tuned model demonstrates superior performance compared to the model trained exclusively on real-world data. }
    \label{fig:forinstance}
\end{figure}

For the instance segmentation task, performance trends similar to those observed in semantic segmentation are evident, as demonstrated in Table \ref{tab:instance_results}. When trained exclusively on Boreal3D, the TreeLearn model failed to converge, likely due to its clustering-based methodology that relies on accurately predicting offset spaces, where each point is trained to move toward its ground-truth center. The domain gap between synthetic and real-world data likely compromised the accuracy of these predicted offsets, preventing the model from achieving clustering convergence. In contrast, OneFormer3D demonstrated acceptable performance when trained solely on Boreal3D, achieving an F1-score of 52.5\%.  In the mixed training scenario, where Boreal3D is combined with real-world data, TreeLearn continued to underperform, while OneFormer3D achieved promising results, surpassing the performance of models trained exclusively on real-world data by 7.3\% in F1-score. Notably, fine-tuning pre-trained models with real-world data yielded significant improvements for both models. Specifically, TreeLearn achieved an F1-score of 82.3\%, highlighting the substantial enhancement in segmentation capability enabled by the fine-tuning approach. Figure \ref{fig:forinstance} provides a visual representation of the instance segmentation results obtained using the TreeLearn method. 

Overall, the experimental results for both semantic and instance segmentation demonstrate that directly transferring models trained on Boreal3D to real-world datasets leads to reduced performance compared to models trained exclusively on real-world data. However, when Boreal3D is combined with real-world data during training, model performance improves significantly, with certain architectures surpassing the results achieved by models trained solely on real-world datasets. Most notably, the fine-tuning of pre-trained models using real-world data yields the best performance across all scenarios. These findings underscore the effectiveness of a two-stage approach: leveraging synthetic data for initial pre-training followed by domain-specific fine-tuning with real-world data, which consistently enhances segmentation performance in forest structure analysis.

To further investigate the data efficiency of fine-tuning pre-trained models, we conducted a series of experiments with varying proportions of real-world annotations. Specifically, we randomly sampled 50\%, 20\%, 10\%, and 1\% of the real data for fine-tuning training. For semantic segmentation, we employed PTv3 as the model architecture. The results, presented in Table \ref{tab:semantic_segmentation_proportion}, reveal several key insights: When using 50\% or 20\% of the real labels, the model achieved performance comparable to fine-tuning with the full dataset (100\% of real labels), with the OA decreasing by only 1.2\% and both mACC and mIoU decreasing by less than 10\%. However, when the proportion of real labels was reduced below 10\%, model performance declined significantly, with the mIoU dropping by more than 10\%. For instance segmentation, we utilized the TreeLearn model to evaluate the impact of reduced fine-tuning data. Similar results were obtained. The experimental results in Table \ref{tab:instance_segmentation_proportion} demonstrate that fine-tuning with 50\% and 20\% of the real-world labels resulted in only marginal performance degradation, with F1-scores decreasing by 3\% and 4\%, respectively. However, when the proportion of labels was reduced to 10\%, a more pronounced decline in performance was observed: the F1-score dropped by 8.5\%, accompanied by a 13.5\% increase in commission error. Additional experiment results exploring various data selection strategies for fine-tuning are provided in the supplementary materials.

These findings suggest that fine-tuning pre-trained models can achieve near-optimal performance with as little as 20\% of the real-world annotations. This underscores the remarkable data efficiency of the proposed approach, which significantly reduces the annotation burden while maintaining high segmentation accuracy in forest structure analysis.

\begin{table}[t]
\centering
    \resizebox{0.35\linewidth}{!}{
        \begin{tabular}{cccccc}
        \toprule
            Model & Proportion & OA (\%) & mACC (\%) & mIoU (\%) \\
        \midrule
        PTv3        & 100\% & 93.61 & 91.30 & 82.40   \\
                    & 50\%  & 93.63 & 88.93 & 81.50 \\
                    & 20\%  & 92.40 & 82.11 & 76.50 \\
                    & 10\%  & 90.78 & 77.04 & 70.95 \\
                    & 1\%   & 73.39 & 54.43 & 37.52 \\
        \bottomrule
        \end{tabular}
    }
        \caption{
            Semantic segmentation results on real data under supervision with different proportions of labeled data.
        }
        \label{tab:semantic_segmentation_proportion}
\end{table}

\begin{table}[t]
    \centering
    \resizebox{0.6\linewidth}{!}{
    \begin{tabular}{cccccc}
    \toprule
        Model & Proportion & Completeness (\%)  & Omission error (\%) & Commission error (\%) & F1-score (\%) \\
    \midrule
    TreeLearn & 100\% & 81.4 & 18.6 & 15.9 & 82.3 \\
                & 50\%  & 79.6 & 20.4 & 20.1 & 79.3 \\
                & 20\%  & 80.1 & 19.9 & 22.2 & 78.3 \\
                & 10\%  & 79.0 & 21.0 & 29.4 & 73.8 \\
    \bottomrule
    \end{tabular}
    }
    \caption{
        Instance segmentation results on real data under supervision with different proportions of labeled data.
    }
    \label{tab:instance_segmentation_proportion}
\end{table}

\subsubsection{Boreal3D for Cross-platform Forest Point Cloud Analysis}
To comprehensively assess the generalizability and potential of Boreal3D across diverse data acquisition platforms, we conducted additional experiments using datasets from ALS, MLS, and TLS platforms. These experiments aim to evaluate the cross-platform applicability of Boreal3D and its effectiveness in addressing platform-specific challenges in forest structure analysis.

\begin{figure}
	\centering
	\includegraphics[width=0.3\linewidth]{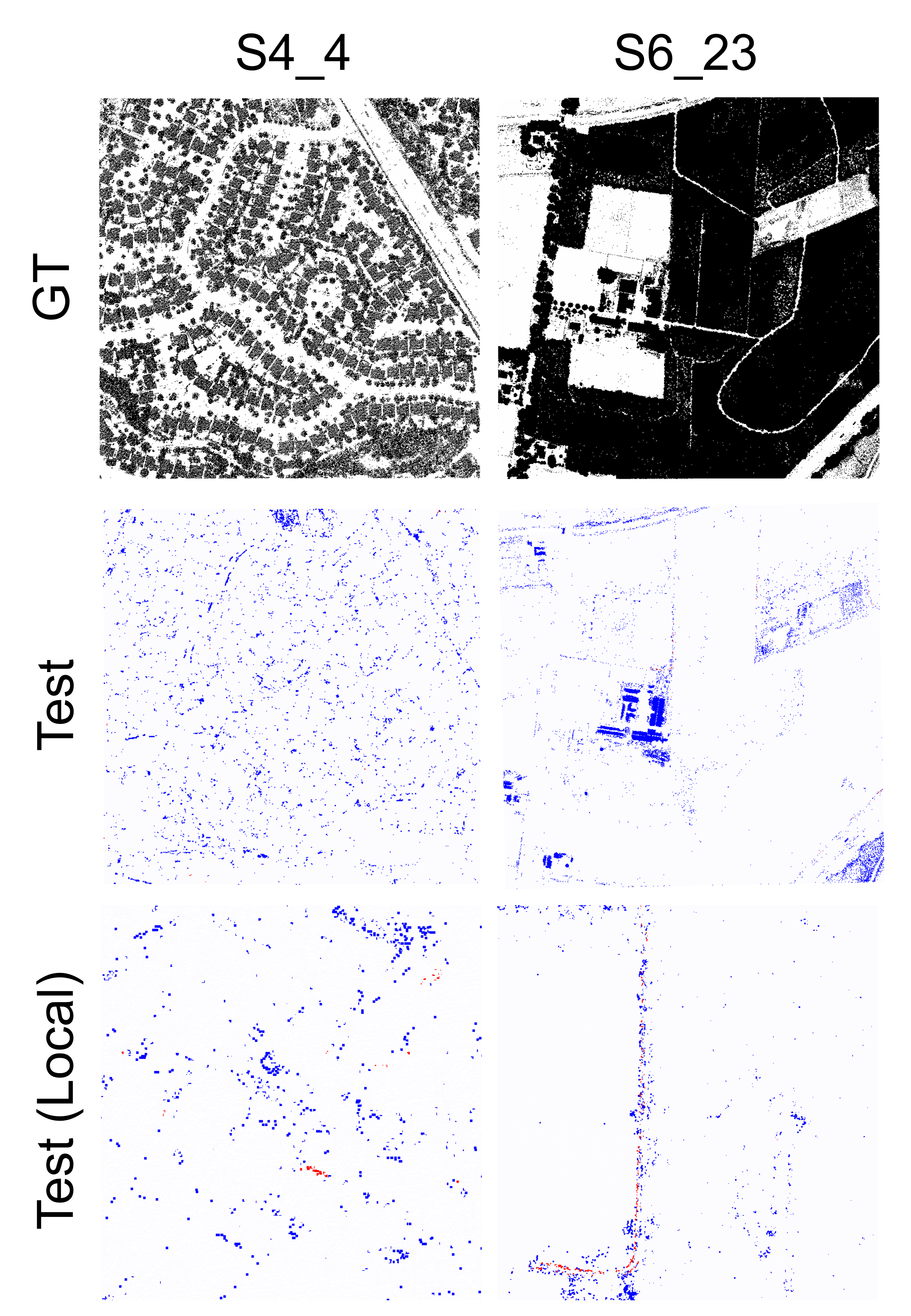}
	\caption{Qualitative results of ground filtering for two test plots in OpenGF. The top row shows the ground truth. Red points indicate non-ground points that were misclassified as ground, while blue points represent ground points that were erroneously classified as non-ground. }
    \label{fig:als}
\end{figure}

For ALS, a groung filtering model was pre-trained with synthetic ALS point clouds in Boreal3D. Fine-tuned with real data from OpenGF, the fine-tuned model was tested on two plots named S4\_4 and S6\_23. Results were compared with Terrain-Net \citep{rs14225798}, which is a dedicated deep learning model for ground filtering in forest environments. Table \ref{tab:ALS_results} shows that the segmentation accuracy of S4\_4 and S6\_23 exceeded that of the original Terrain-Net. The S6\_23 plot was an urban enviroments that is very different from the terrain constructed in this paper, resulting in the worst pre-training performance. However, the substantial performance improvement after fine-tuning further demonstrated the enhancement effect of synthetic data on the model over real data. Figure \ref{fig:als} intuitively shows the prediction results of the fine-tuned model on two test plots. The number of ground points that the model misclassifies as non-ground points (blue) is greater than the number of non-ground points that the model misclassifies as ground points (red).

For ALS platform, a ground filtering model was pre-trained using synthetic ALS point clouds from Boreal3D. The model was subsequently fine-tuned with real-world data from the OpenGF dataset and evaluated on two test plots: S4\_4 and S6\_23. The results were compared against Terrain-Net \citep{rs14225798}, a state-of-the-art deep learning model specifically designed for ground filtering in forest environments. As shown in Table \ref{tab:ALS_results}, the fine-tuned model achieved superior segmentation accuracy on both S4\_4 and S6\_23 plots compared to the original Terrain-Net. Notably, the S6\_23 plot, characterized by an urban environment significantly different from the terrain represented in Boreal3D, exhibited the weakest pre-training performance. However, the substantial performance gains observed after fine-tuning further underscore the effectiveness of leveraging synthetic data for model enhancement. Figure \ref{fig:als} provides a visual comparison of the fine-tuned model's predictions on the two test plots. The visualization reveals that the number of ground points misclassified as non-ground points (indicated in blue) exceeds the number of non-ground points misclassified as ground points (indicated in red), offering insights into the model's error patterns and areas for potential improvement.

\begin{table}[t]
    \centering
    \resizebox{0.6\linewidth}{!}{
    \begin{tabular}{cccccccc}
    \toprule
    Plot ID & Model & \multicolumn{2}{c}{Ground} & \multicolumn{2}{c}{Non-Ground} & \multirow{2}{*}{OA (\%)} & \multirow{2}{*}{Kappa (\%)}  \\
         & & IoU (\%) & F1-score (\%) & IoU (\%) & F1-score (\%) \\
    \midrule
    S4\_4 & Terrain-Net & 86.0  & 92.4  & 82.3 & 90.3 & 91.5  & 82.9  \\
          & Pre-train    & 85.0  & 89.0 & 90.6 & 95.0 & 93.9 & 87.0 \\
          & Fine-tune    & 93.8 & 97.0 & 93.5 & 96.7 & 96.7 & 93.5 \\ 
    \midrule
    S6\_23 & Terrain-Net & 92.8  & 96.3 & 94.6 & 97.2 & 96.6 & 93.5 \\
           & Pre-train    & 14.5 & 11.1 & 80.7 & 89.3 & 81.3 & 18.8 \\
           & Fine-tune    & 94.7 & 96.2 & 96.9 & 98.4 & 98.0 & 95.7 \\
    \bottomrule
    \end{tabular}
}
    \caption{
        Comparative experimental results on the ground filtering task for ALS data.
    }
    \label{tab:ALS_results}
\end{table}

\begin{figure}
	\centering
	\includegraphics[width=0.9\linewidth]{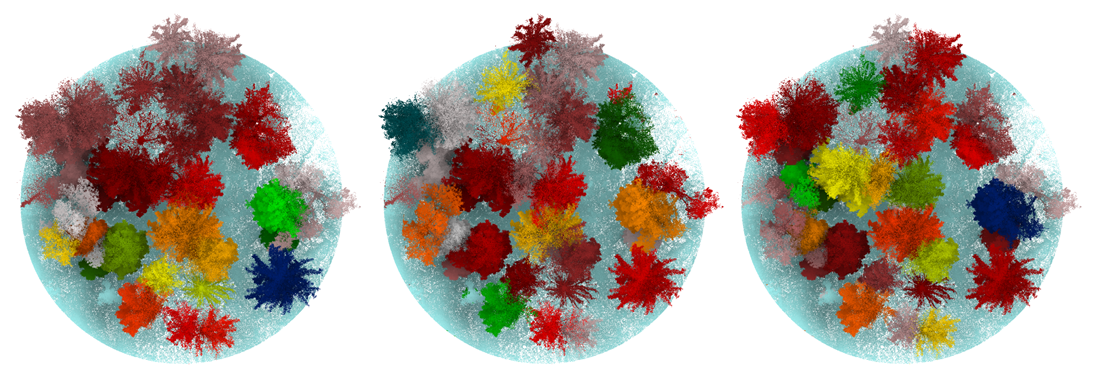}
	\caption{Instance segmentation results obtained using the TreeISO and fine-tuned TreeLearn methods on the MLS test set. From left to right: (1) the ground truth (GT) annotations, (2) the segmentation results generated by TreeISO, and (3) the results produced by fine-tuned TreeLearn. }
    \label{fig:mls}
\end{figure}

For MLS data, we pre-trained the model using 100 synthetic MLS plots in Boreal3D and fine-tuned it with real-world data from the Point2Tree (P2T) dataset \citep{rs15153737}. We conducted instance segmentation experiments using two methods: TreeISO \citep{rs14236116} and TreeLearn \citep{henrich2024treelearn}, reporting the averaged test results on four randomly selected plots. As shown in Table \ref{tab:MLS_results}, the fine-tuned TreeLearn model demonstrated significant performance improvements compared to the TreeISO approach. Notably, TreeISO exhibited a high omission error rate, indicating its limitations in accurately detecting small trees. Figure \ref{fig:mls} provides a visual comparison of the segmentation results, highlighting TreeISO's inability to effectively distinguish small, overlapping trees. In contrast, TreeLearn, particularly after fine-tuning, showed enhanced capability in resolving such challenging cases, further validating the effectiveness of the proposed approach for MLS-based forest structure analysis.

\begin{table}[t]
    \centering
    \resizebox{0.7\linewidth}{!}{
    \begin{tabular}{ccccccc}
    \toprule
    Method &  Completeness (\%) & Omission error (\%) &  Commission error (\%) & F1-score (\%) \\
    \midrule
    TreeISO  & 62.67 & 37.33 & 12.41 & 72.29 \\
    Fine-tuned TreeLearn & 89.03 & 10.97 & 13.61 & 87.31 \\
    \bottomrule
    \end{tabular}
}
    \caption{
        Comparative experimental results on the instance segmentation task for MLS data.
    }
    \label{tab:MLS_results}
\end{table}

\begin{figure}
	\centering
	\includegraphics[width=0.9\linewidth]{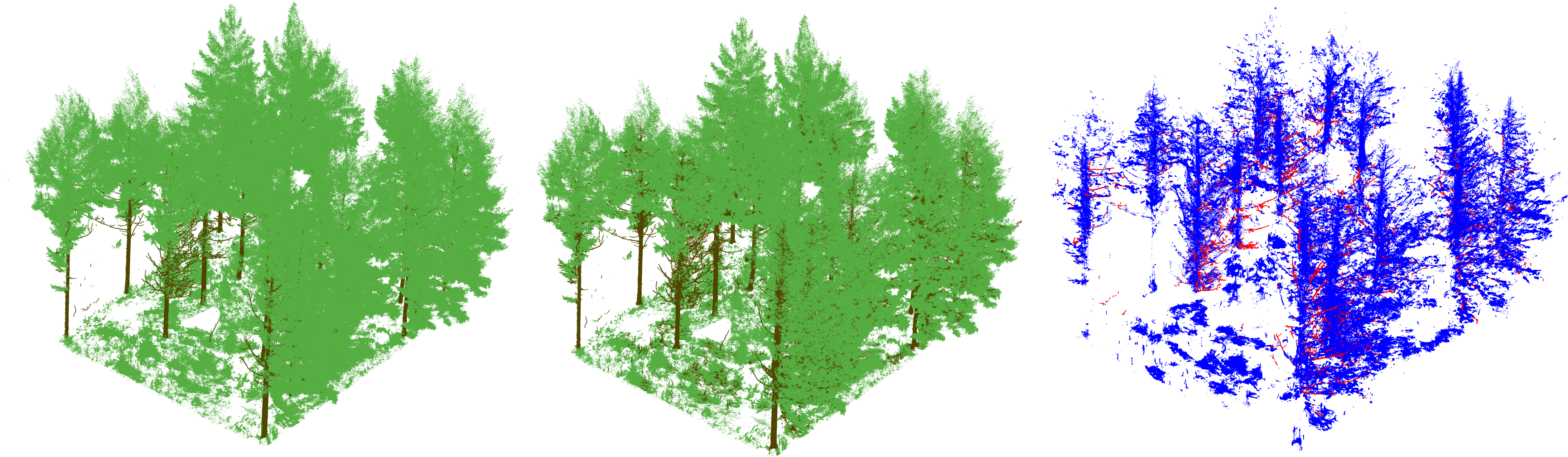}
	\caption{Visualization of wood-leaf separation Results. From left to right: (1) the ground truth annotations, (2) the segmentation results generated by the fine-tuned model, and (3) the misclassified points. In the visualization, red points represent wood points erroneously classified as leaf, while blue points indicate leaf points mistakenly classified as wood. }
    \label{fig:TLS}
\end{figure}

For the TLS evaluation, we pre-trained the model using synthetic TLS data from 100 plots in Boreal3D and fine-tuned it on real-world TLS forest scenes. Notably, we implemented two distinct fine-tuning strategies for TLS data: (1) fine-tuning by randomly sampling a fixed number of points from the real plot, and (2) fine-tuning by splitting the original plot into smaller segments. As demonstrated in Table \ref{tab:TLS_results}, both fine-tuning approaches significantly enhanced model performance, with each method offering unique advantages in addressing the challenges of TLS-based forest structure analysis. Figure \ref{fig:TLS} displays the results of leaf-wood separation. These results further validate the effectiveness of synthetic data pre-training combined with domain-specific fine-tuning for improving segmentation accuracy in diverse LiDAR applications.

\begin{table}[t]
    \centering
    \resizebox{0.55\linewidth}{!}{
    \begin{tabular}{cccc}
    \toprule
    Plot & Method & OA (\%) & Kappa (\%) \\
    \midrule
    White Birch & \cite{BRODU2012121} & 91.11 & 64.93 \\
                & \cite{tls-wood-leaf} & 94.61 & 77.11 \\
                & Pre-train & 88.67 & 51.78 \\
                & Pre-train + Fine-tune1 & 98.05 & 86.74 \\
                & Pre-train + Fine-tune2 & 98.28 & 86.32 \\
    Dahurian Larch & \cite{BRODU2012121} & 90.68 & 73.02 \\
                & \cite{tls-wood-leaf} & 95.26 & 85.90 \\
                & Pre-train & 86.74 & 44.68 \\
                & Pre-train + Fine-tune1 & 96.64 & 77.42 \\
                & Pre-train + Fine-tune2 & 96.72 & 79.28 \\
    Chinese scholar tree & \cite{BRODU2012121} & 89.50 & 65.02 \\
                & \cite{tls-wood-leaf} & 94.07 & 77.38 \\
                & Pre-train & 86.42 & 53.34 \\
                & Pret-rain + Fine-tune1 & 96.63 & 82.88 \\
                & Pret-rain + Fine-tune2 & 96.66 & 82.97 \\

    \bottomrule
    \end{tabular}
}
    \caption{
        Comparative experimental results on the leaf-wood separation task for TLS data.
    }
    \label{tab:TLS_results}
\end{table}
\section{Discussion}
\label{sec:Discussion}
\subsection{Sim2Real for Forests}
Digital simulation is an effective way to solve the problem that it is difficult to obtain labels for point cloud data in forest scenes. In recent years, a large number of research works have verified the effectiveness of simulation in various fields \citep{tao2018digital}. For example, in the field of engineering, digital twin technology with digital simulation as the core is used to simulate and interpret various special situations in order to make safety warnings and response plans in advance \citep{sleiti2022digital}. In the field of smart cities, digital twin technology is widely used to explore infrastructure planning, predict and alleviate traffic congestion and environmental degradation. In short, the core of simulation is to build a digital scene model that can represent the real environment \citep{white2021digital}. However, constructing digital twin models for complex forest scenes remains challenging due to extensive occlusions and intricate details, which require vast amounts of high-quality, densely annotated data to accurately reflect real-world complexity. While recent studies have made progress, current capabilities fall short of meeting the demands for modeling large-scale forest ecosystems \citep{qiu2023forest}. Existing data and annotation methods are insufficient, highlighting the need for innovations in data acquisition, annotation techniques, and modeling frameworks to achieve scalable and realistic forest digital twins.

Relying solely on digital twins is insufficient for advancing large-scale forest scene understanding. This paper introduces a novel approach based on the concept of digital cousins, constructing a digital model derived from real forest scenes. The core innovation lies in developing a fully automated, highly scalable, and versatile simulation process capable of generating fine-grained forest scenes of any scale and type (e.g., tree species, terrain). This significantly enhances the practical utility of virtual forests. Our experiments demonstrate that, when combined with LiDAR simulation, this framework can generate multi-platform forest point clouds, facilitating fine-grained 3D forest structure analysis. Additionally, synthetic data offers the advantage of 100\% accurate annotations, unlike human annotations which are prone to errors. For instance, as shown in Figure \ref{fig:TLS}, many points misclassified as leaf by human annotators are actually wood, highlighting the importance of annotation reliability. Compared with similar work at this stage, for example, \citet{GAO2024133} used digital simulation technology to synthesize RGB images of rice, and combined with frame images for the development of rice semantic segmentation models at different spatial resolutions. \citet{LIU2023113832} used the Discrete Anisotropic Radiative Transfer Model (DART) to simulate the Bidirectional Reflectance Factor (BRF) of 3D explicit forest scenes, further studying and developing more accurate remote sensing products and promoting the understanding of forest radiation transfer processes. These similar works have also demonstrated that digital simulation technology can promote researchers to explore the development and application of multi-domain and multi-modal data.

Building on the rapid advancements in 3D generation, future work could explore the integration of generative models to automate the positioning, layout, and parameterization of trees within forest scenes. Such innovations could significantly enhance forest scene understanding, enabling the creation of diverse and realistic virtual forest environments that better support ecological research and applications.

\subsection{Advantages and Benefits of Boreal3D}
In recent years, the remote sensing and forestry communities have increasingly turned to deep learning techniques for processing point cloud data, aiming to enhance the analysis of complex 3D forest structures \citep{XIANG2024114078}. However, deep learning models require extensive training data, prompting researchers to develop and release several forest point cloud datasets. Despite these efforts, significant challenges remain: the intricate structure of forest point clouds necessitates labor-intensive and time-consuming semantic and instance annotations, while the high heterogeneity of forest environments makes it difficult for manually annotated datasets to encompass diverse forest types. These limitations have hindered the ability of real-world forest datasets to meet current research demands, thereby constraining the development and application of advanced vision and artificial intelligence technologies, such as deep learning, in forest point cloud analysis.

Our work addresses these challenges by providing an efficient and scalable solution. Leveraging synthetic data generation, our framework enables the creation of annotated datasets without scale restrictions and allows for the simulation of virtually any forest scene. As demonstrated by the experiments in this study, synthetic forest point cloud data not only aids models in understanding real-world forest scenes but also facilitates deeper analysis of complex forest structures. This breakthrough is expected to significantly accelerate progress in the field, overcoming the limitations of traditional datasets and fostering innovation in forest structure analysis and ecological monitoring.

Moreover, Boreal3D offers a unique advantage in its multi-platform coverage capability. The proposed forest Digital Cousins and Sim2Real framework enables the generation of simulated data across diverse platforms, including but not limited to ALS, ULS, MLS, and TLS. This comprehensive dataset supports a wide range of tasks in forest scene understanding, such as point cloud registration, semantic segmentation, and 3D modeling. Currently, most research in this field relies on data from a single platform due to the prohibitive costs associated with acquiring multi-platform datasets and the challenges of achieving comprehensive coverage for individual tasks. The multi-platform nature of Boreal3D addresses these limitations, providing researchers with a versatile and cost-effective resource. This innovation is expected to catalyze future research in emerging areas such as multi-platform integration and multi-task learning, potentially revolutionizing the field of forest structure analysis and ecological monitoring.

Last but not least, beyond segmentation tasks, Boreal3D provides precise ground truth values for key forest structural parameters, which are critical for advanced forest research. In traditional field studies, obtaining accurate measurements of parameters such as Leaf Area Index (LAI), volume (biomass), and other essential structural metrics often requires destructive sampling methods, which are both labor-intensive and ecologically disruptive. Our framework addresses this limitation by offering error-free, annotated ground truth values for these parameters through synthetic data generation. While recent studies have begun exploring the use of synthetic data for validating predictions of LAI and biomass \citep{liu2021comparative, cpad061, yu2024evaluating}, these efforts remain in their nascent stages, with limited scope and development. The comprehensive methods and datasets we provide are poised to significantly advance research in this domain, enabling more accurate, non-destructive estimation of forest structural parameters and fostering innovation in ecological monitoring and forest management.

\subsection{Challenges and Future Works}
Despite the promising potential of the proposed framework, several challenges remain to be addressed. First, understanding forest scenes requires models to process point clouds at the plot level, which involves handling massive amounts of data simultaneously. This leads to high computational resource demands, making it difficult to process large-scale forest scenes without access to high-performance computing infrastructure. Second, as evident in the experimental results, a noticeable gap persists between synthetic and real-world data. This discrepancy arises because the structural complexity of real forest scenes cannot be fully captured by synthetic data, even with advanced modeling techniques. While synthetic data provides a valuable approximation, it may not account for all structural parameters and natural variations present in real forests. Finally, the redundancy and heterogeneity of data across different platforms (e.g., ALS, MLS, TLS) pose significant challenges. Current algorithms and models are not yet capable of efficiently integrating and leveraging multi-platform data, limiting their ability to fully exploit the complementary characteristics of each platform. Addressing these limitations will require advancements in data fusion techniques, computational efficiency, and model architectures tailored to the unique challenges of multi-platform forest scene analysis.

In recent years, advanced training methodologies such as domain adaptation, multi-platform joint learning, and multi-task joint learning have garnered significant attention across various domains. However, their potential remains largely untapped in the context of complex forest scenes, where these techniques could offer transformative solutions. Future research will focus on two key directions: First, we aim to further bridge the gap between synthetic and real-world data by refining simulation techniques and leveraging domain adaptation methods. This will enhance the ability of synthetic data to improve models' intelligent parsing capabilities for real-world forest scenes. Second, while our work addresses the critical issue of data scarcity in multi-platform forest scenarios, we plan to explore the integration of multi-platform joint learning and multi-task joint learning frameworks. These approaches will enable more comprehensive analysis and verification of forest structural parameters, such as biomass estimation, canopy characterization, and ecological monitoring, ultimately advancing the field of forest informatics and sustainable management.
\section{Conclusion}
\label{sec:Conclusion}
The understanding of large-scale, complex forest scenes has long been hindered by the scarcity of high-quality data and annotations. Addressing this critical challenge, this study introduces an automated point cloud data synthesis framework grounded in the innovative concepts of Digital Cousins and Simulation-to-Reality (Sim2Real). This framework facilitates the generation of point cloud data at any scale and for any platform, enriched with error-free semantic, instance, and structural attribute annotations. By bridging the gap between synthetic and real-world data, this work aims to significantly enhance the capability of understanding fine-grained 3D forest structures in real-world scenarios, thereby advancing research in forest ecology, management, and conservation.

Our experimental results highlight three key contributions: First, synthetic data significantly enhances model performance on real-world datasets across multiple tasks, demonstrating its potential to bridge the simulation-to-reality gap. Second, pre-training on synthetic datasets, followed by fine-tuning with minimal real-world annotations (as little as 20\% of the data), achieves performance comparable to models trained exclusively on real-world data, underscoring the data efficiency of our approach. Third, comprehensive evaluations across multiple platforms (e.g., ALS, MLS, TLS) and diverse tasks validate the versatility and effectiveness of synthetic datasets in advancing forest scene analysis, paving the way for scalable and robust solutions in ecological research and management.

The proposed framework tackles the key problem of data scarcity in forest scene understanding by offering a strong and scalable solution. It not only supports progress in complex scene analysis and ecological studies but also creates opportunities for research in multi-platform and multi-task learning. By allowing the creation of diverse, high-quality synthetic datasets with accurate labels, this work has the potential to transform forest research, driving advances in sustainable forest management, biodiversity tracking, and climate change efforts.

\section{Supplementary Materials}
\label{sec:supp}
In this section, we present additional analysis of the experimental results on the FOR-instance dataset, along with an in-depth exploration of fine-tuning strategies. This analysis aims to provide deeper insights into the effectiveness of synthetic data pre-training and the optimization of fine-tuning approaches for enhancing model performance in real-world forest scene understanding tasks.

\subsection{Results on Subplots of FOR-instance Dataset}
The FOR-instance dataset comprises five distinct subsets, each representing different forest conditions. In this section, we analyze the semantic segmentation performance of the fine-tuned PTv3 model across these subsets and individual semantic categories. Table \ref{tab:semantic_segmentation_sub} summarizes the evaluation metrics for the fine-tuned PTv3 method on each subset. As shown in the table, the model's performance varies significantly across subsets, reflecting differences in tree densities, spatial distributions, and ecological characteristics within each plot.

Furthermore, the performance for individual semantic categories also exhibits considerable variation. For instance, in the CULS subplot, the model entirely missed the understory category, whereas in the NIBIO subplot, the understory achieved an Intersection over Union (IoU) of 95.29\%. These variations underscore the challenges posed by forest heterogeneity and highlight the need for robust models capable of adapting to diverse forest conditions and accurately segmenting all semantic categories.

\begin{table}[t]
    \centering
    \resizebox{0.6\linewidth}{!}{
    \begin{tabular}{ccccccccc}
    \toprule
        Model & Test & OA (\%) & mACC (\%) & mIoU (\%) & \multicolumn{4}{c}{IoU (\%)} \\
               &      &    &      &       & Understory& Terrain & Leaf & Wood\\ 
    \midrule
    PTv3 & ALL    & 93.61 & 91.30 & 82.40 & 90.55 & 81.56 & 92.95 & 64.75 \\
         & CULS   & 96.29 & 71.26 & 66.13 & 0     &  99.76 & 93.82 & 70.95 \\
         & NIBIO  & 94.63 & 74.54 & 68.44 & 95.29 & 15.92 & 93.45 & 69.12 \\
         & RMIT   & 81.58 & 65.71 & 50.69 & 1.79  & 85.05 & 74.60 & 41.32 \\
         & SCION  & 89.49 & 69.35 & 52.70 & 0.30  & 62.77 & 92.64 & 55.36 \\
         & TUWIEN & 88.33 & 68.81 & 55.99 & 5.26  & 92.39 & 86.18 & 40.13 \\
    \bottomrule
    \end{tabular}
}
    \caption{
        Performance of the fine-tuned PTv3 model across various subplots in the For-instance dataset.
    }
    \label{tab:semantic_segmentation_sub}
\end{table}

\subsection{Results on Different Fine-tuning Strategies}

In addition to proportionally reducing the amount of real-world labeled data, we also investigated model performance by controlling the number of labeled plots used during fine-tuning. Using PTv3 as an example, we fine-tuned the model with labels from individual plots to evaluate its impact. Table \ref{tab:semantic_segmentation_plot} presents the performance of PTv3 under varying numbers of plot-based supervision.

The results reveal that fine-tuning with labels from individual plots leads to inconsistent model performance, as the choice of specific plots significantly influences the outcomes. This instability highlights the variability in forest characteristics across different plots and underscores the importance of using diverse and representative training data to ensure robust model generalization.

\begin{table}[t]
    \centering
    \resizebox{0.3\linewidth}{!}{   
    \begin{tabular}{cccccc}
    \toprule
        Model & Plot Count & OA (\%) & mACC (\%) & mIoU (\%) \\
    \midrule
    PTv3        & 1*  & 87.78 & 73.20 & 62.64 \\
                & 1  & 78.97 & 62.44 & 47.51 \\
                & 2  & 83.64 & 72.47 & 57.96 \\
                & 3  & 80.83 & 69.95 & 53.81 \\
                & 4  & 88.23 & 76.20 & 65.19 \\
                & 5  & 88.61 & 81.17 & 68.41 \\
                & ALL & 93.61 & 91.3 & 82.4 \\
    \bottomrule
    \end{tabular}
}
    \caption{
       Semantic segmentation performance with varying plot-based fine-tuning supervision. The notations '1*' and '1' represent two distinct randomly selected plots used for supervision.
    }
    \label{tab:semantic_segmentation_plot}
\end{table}

\section*{Data availability}
Boreal3D is available at: \url{https://boreal3d.github.io}
\section*{Acknowledgments}
This work is supported by the National Natural Science Foundation of China (No. 42101330), the National Key Research and Development Program of China (No. 2021YFF0704600), and the Key Research and Development Program of Shaanxi Province (No. 2023-YBSF-452).

\bibliographystyle{elsarticle-harv}
\bibliography{Simulation-to-Reality}

\end{document}